\documentclass[pdflatex, sn-mathphys-num]{sn-jnl}
\usepackage[utf8]{inputenc}
\usepackage{adjustbox}
\usepackage[vlined,ruled,linesnumbered,english,onelanguage]{algorithm2e}
\usepackage{amsmath}
\usepackage{amssymb}
\usepackage{booktabs, caption}
\usepackage{floatrow}
\usepackage{makecell}
\usepackage{multirow, multicol}
\usepackage{moresize}
\usepackage{fontawesome5}
\usepackage{hyperref}
\usepackage[capitalise,nameinlink,noabbrev]{cleveref}
\usepackage{nomencl}
\usepackage[dvipsnames]{xcolor}
 \usepackage{bookmark}

\makenomenclature
\def\arraystretch{1.4}
\newfloatcommand{capbtabbox}{table}[][\FBwidth]

\begin{document}

\title{MetaChest: Generalized few-shot learning of pathologies from chest X-rays}

\author*[1,2]{\fnm{Berenice} \sur{Montalvo-Lezama}}\email{bereml@turing.iimas.unam.mx}

\author[2]{\fnm{Gibran} \sur{Fuentes-Pineda}}\email{gibranfp@unam.mx}

\affil*[1]{\orgdiv{Posgrado en Ciencia e Ingeniería de la Computación}}

\affil[2]{\orgdiv{Instituto de Investigaciones en Matemáticas Aplicadas y en Sistemas}, \orgname{Universidad Nacional Autónoma de México}, \orgaddress{\street{Circuito Escolar s/n, Ciudad Universitaria}, \city{Coyoacán}, \postcode{04510}, \state{CDMX}, \country{Mexico}}}

\abstract{
The limited availability of annotated data presents a major challenge for applying deep learning methods to medical image analysis. 
Few-shot learning methods aim to recognize new classes from only a small number of labeled examples. 
These methods are typically studied under the standard few-shot learning setting, where all classes in a task are new.
However, medical applications such as pathology classification from chest X-rays often require learning new classes while simultaneously leveraging knowledge of previously known ones, a scenario more closely aligned with generalized few-shot classification. 
Despite its practical relevance, few-shot learning has been scarcely studied in this context.
In this work, we present MetaChest, a large-scale dataset of 479,215 chest X-rays collected from four public databases. 
MetaChest includes a meta-set partition specifically designed for standard few-shot classification, as well as an algorithm for generating multi-label episodes. We conduct extensive experiments evaluating both a standard transfer learning approach and an extension of ProtoNet across a wide range of few-shot multi-label classification tasks. 
Our results demonstrate that increasing the number of classes per episode and the number of training examples per class improves classification performance. Notably, the transfer learning approach consistently outperforms the ProtoNet extension, despite not being tailored for few-shot learning. 
We also show that higher-resolution images improve accuracy at the cost of additional computation, while efficient model architectures achieve comparable performance to larger models with significantly reduced resource requirements.
}


\maketitle
\section{Introduction}

In recent decades, deep learning has revolutionized medical image analysis, particularly in the field of radiology \citep{tfmedical, chextransfer2021, cherti2022}. Deep neural networks have enabled the processing of large volumes of radiological data, the extraction of complex features, and the development of models that can enhance the accuracy of medical diagnoses. 
Despite these advances, a major challenge arises when only limited annotated data are available, as deep learning models typically require large amounts of labeled data to achieve strong performance.
This issue is especially relevant to tasks such as pathology classification in chest X-rays, where labeled data can be scarce and difficult to obtain.
To address this limitation, early research has explored the use of the standard few-shot classification (SFSC) paradigm, which aims to train models capable of generalizing to new classes using only a few labeled examples per class.
However, this paradigm differs significantly from the way pathologies manifest in practice. 
From a clinical perspective, the objective is not merely to classify entirely new disease categories, but rather to distinguish between a combination of known pathologies and previously unseen ones. 
This highlights the need for approaches that are more closely aligned with the complexities of clinical settings.

This work aims to investigate the factors that influence the training of pathology classification models under a formulation that more closely reflects clinical scenarios.
In particular, we study how different task instance configurations within the generalized few-shot learning (GFSL) paradigm affect model performance.
In addition, we compare two training methods derived from the standard transfer learning and standard few-shot learning paradigms, evaluating them on tasks with a generalized few-shot learning formulation.
Finally, we analyze the impact of image resolution and neural network architecture on classification performance.

To address these objectives, this paper makes the following contributions:
\begin{itemize}
    \item We introduce MetaChest, a dataset comprising 479,215 chest X-ray images collected from four public databases, along with a meta-set partition specifically designed for standard few-shot classification.
    \item We provide an algorithm to generate multi-label episodes, enabling few-shot learning in multi-label settings.
    \item We propose ProtoNet-ML, an extension of ProtoNet for multi-label classification tasks.
    \item We conduct a comprehensive comparison of two methods, one based on standard transfer learning and the other on standard few-shot learning, across a wide range of tasks with varying complexity.
    \item We analyze the influence of image resolution and model architecture on pathology classification performance.
\end{itemize}

The rest of this article is organized as follows. 
\cref{sec:relwork} reviews prior work on pathology classification in chest X-rays using deep learning, including approaches based on transfer learning and meta-learning.
\cref{sec:MetaChest} introduces the MetaChest dataset, detailing its pathology distribution and a partitioning scheme suitable for standard few-shot classification.
\cref{sec:transfer} outlines the key differences between standard transfer learning, standard few-shot classification, and generalized few-shot classification.
\cref{sec:episodes} proposes using generalized few-shot classification to generate tasks that more closely resemble clinical practice in pathology classification.
It also introduces a multi-label episode generation algorithm and describes the two classification methods used in this work: BatchBased and ProtoNet-ML.
\cref{sec:experimental} details the experimental setup and evaluation methodology, and presents and analyzes the experimental results.
Finally, \cref{sec:conclu} summarizes the conclusions and outlines directions for future research.

\section{Related work}
\label{sec:relwork}
In this section, we review related works on chest X-ray classification using deep learning techniques. 
We also discuss relevant transfer learning and meta-learning approaches, as well as their applications to the medical image domain.

\subsection{Deep learning for chest X-ray classification}

Deep neural networks, coupled with large-scale datasets, have enabled significant progress in several computer vision fields. 
Over the past few years, in an effort to take advantage of deep neural networks, datasets of increasingly specialized domains have been made publicly available. 
For instance, in the medical domain, multiple chest X-ray datasets have been introduced, such as CheXpert~\citep{CheXpert2019}, Chest X-ray8~\citep{chestX-ray82017}, Chest X-ray14~\citep{chestX-ray82017}, MIMIC~\citep{MIMIC-CXR2019}, MIMIC-CXR-JPG~\citep{mimicjpg}, OpenI~\citep{openI}, and PadChest~\citep{PadChest}.
These datasets have been fundamental in the development of deep learning models for chest X-ray analysis and generation tasks.
In contrast to ImageNet~\citep{ImageNet}, the scale of these datasets is at least one order of magnitude smaller.
In addition, the distributions of these datasets are greatly heterogeneous, e.g. the number and kind of pathologies, the class imbalance, the collection and labeling procedure, the quality of the images, and the patient population.   

With the introduction of these datasets, several studies addressing pathology classification from chest X-rays using deep learning started to appear. 
For binary classification (i.e. presence or absence), \citet{tuberculosiscls} focused on tuberculosis identification, whereas \citet{pneumoniadetection} targeted pneumonia identification.
Since the X-ray of a patient may exhibit signs of multiple diseases, the identification of pathologies from chest X-rays has often been formulated as a multi-label classification problem.    
For instance, \citet{comparisonmodels2019} used a ResNet-50~\citep{resnet} architecture to classify fourteen pathologies in the ChestX-ray14 dataset, where each X-ray could be assigned to more than one pathology. 
Similarly, \citet{CheXpert2019} compared various ConvNet architectures for multi-label chest X-ray classification using CheXpert, finding that DenseNet121 outperformed
ResNet152~\citep{resnet}, Inception-v4~\citep{inceptionv4}, and SE-ResNeXt101~\citep{se-resnex}.

\subsection{Standard transfer learning}
\label{subsec:transfer}

Transfer learning (TL) is a cornerstone of deep learning for image analysis, since it can reduce the amount of data and the computational resources required to train a model for a target task by leveraging the representations learned from one or multiple source tasks.   
In TL, multiple strategies to adapt the representations from the source task to the target task have been proposed.
In practice, the most widely used transfer strategy has been standard transfer learning, which consists of pre-training models using a conventional batch-based training (as opposed to other training schemes).
Specifically, ImageNet pre-training has been a standard practice for a wide variety of natural image tasks, including classification~\citep{efficientnet, bit2020}, segmentation~\citep{r-cnn,yolo} and object detection~\citep{fcn}.

Due to the widespread use of ImageNet in practice, multiple works have studied the transferability of the learned representations from ImageNet to other natural image tasks~\citep{featuresoff-the-shelf, Yosinski2014, kornblith2019better, zhou2021convnets}.
Surprisingly, even though some studies have suggested that the source and target datasets must be closely related for an effective knowledge transfer~\citep{Yosinski2014}, ImageNet pre-training has been used with seeming success for wildly dissimilar image domains (e.g. medical images~\citep{chestX-ray82017, transfusion2019, chextransfer2021}).
In contrast, there are transferability studies in specific domains where ImageNet pre-training has not provided any improvement over random initialization~\citep{transfusion2019, tfmedical, cherti2022}.

Other transferability studies have focused on analyzing the effect of the architecture size and the scale of the training dataset on the effectiveness of STL.
In intra-domain scenarios, where the source and target datasets are closely related, studies have been mainly focused on natural image datasets. 
For example, \citet{bit2020} and \citet{scalinvisiontsf} analyzed how the pre-training dataset size and the architecture depth influence knowledge transfer when both the source and target datasets are composed of natural images.
The results from these studies have consistently shown better performance with larger architectures and pre-training datasets.  

In inter-domain scenarios, where the source and target datasets belong to different domains (e.g., natural images and chest X-rays), existing studies are scarce, not very systematic and report mixed results. 
\citet{transfusion2019} did not find significant differences on chest X-ray and retinal image classification performance using a ResNet-50 architecture when comparing ImageNet-1k pre-training with random initialization.
\citet{chextransfer2021} studied the effect of ImageNet-1k pre-training on chest X-ray classification performance using ConvNet architectures of different sizes.  
Their results showed a slight performance improvement when using deeper pre-trained architectures. 
On the other hand, \citet{tfmedical} studied the influence of ImageNet-1k, ImageNet-21k, and JFT-300M pre-training on classification performance using ResNets of different sizes. 
The target tasks considered in this study were cancer identification from mammograms, pathology classification from chest X-rays, and skin condition  from dermatological images. 
The results were far from conclusive, observing performance improvements with larger pre-training datasets and architectures only in some target tasks.  
Similarly, \citet{cherti2022} carried out a comparative study of ResNet models pre-trained on ImageNet-1k, ImageNet-21k, and a combination of different chest X-ray datasets for pathology classification.  
They reported small improvements in performance when models were pre-trained on larger source datasets and transferred to larger target datasets.  
However, no performance improvement was observed when transferring for smaller target datasets, regardless of the size of the pre-training dataset.

\subsection{Few-shot classification}

Meta-learning is a transfer learning strategy which aims to generate models that can be quickly adapted to new tasks~\citep{learntolearn}.
As opposed to STL, in meta-learning, new tasks are commonly known as episodes and are typically small with respect to both the number of classes and the number of examples per class. 
The most widely studied problem in meta-learning is few-shot classification, which is a multi-class classification problem where a few examples per class are available for training (typically, 1 or 5).

The earliest works in meta-learning proposed methods for multi-class classification on natural image datasets~\citep{MAML, matching, protonet, relationet, setfeat}, such as MiniImageNet~\citep{matching} and FC100~\citep{meta_transfer}. 
These datasets are reduced versions of ImageNet and CIFAR100 and were created to facilitate episodic training. 

More recent works have applied meta-learning to domain-specific problems, particularly through datasets of various medical imaging modalities. 
For instance, meta-learning methods have been studied in skin disease classification from dermatological images~\citep{metaderm2020}, COVID-19 classification from chest CT scans~\citep{CHEN2021107826}, and cancer classification from histological images~\citep{histopathological2019, shakeri2022fhist}.
Moreover, meta-learning methods have also been used for image segmentation in CT scans, magnetic resonance images~\citep{Tang2021RecurrentMR}, and dermatological images~\citep{KHADKA2022105227}.

\section{MetaChest dataset}
\label{sec:MetaChest}

Over the past decades, several chest X-ray datasets have been collected, which vary in aspects such as number of examples, population of study, labeling strategy, period of time, pathologies, and source institution. 
\autoref{tbl:chest_datasets} shows a comparison of publicly available chest X-ray datasets. 
In general, these datasets exhibit heterogeneous characteristics, ranging from a few thousand to hundreds of thousand of images collected for periods of a few years and up to a few decades. 
One key factor influencing the distribution of pathologies in a dataset is the patient population from which the chest X-rays were obtained.
As can be observed in \autoref{tbl:chest_datasets}, most publicly available datasets were collected from medical institutions in the United States, albeit from different hospitals and regions.
However, there are two datasets from other countries: PadChest, from Hospital San Juan in Spain, and VinDr-CXR, from multiple hospitals in Vietnam. 

Clinical data collection is a complex process involving several tasks which can require a considerable amount of time and resources. 
Data labeling is one of the tasks that can generate greater variability among chest X-ray datasets; the rightmost column in \autoref{tbl:chest_datasets} summarizes the labeling strategy employed by each dataset.
Most datasets derived annotations automatically from radiology reports using natural language processing (NLP) methods, with the exception of PadChest which was annotated by expert radiologists.
The specific strategy and tool used for annotating the chest X-rays directly influences the distribution of labels. 
For instance, MIMIC provides two sets of labels that have different distributions: one generated by NegBio~\citep{negbio} and another by CheXpert~\cite{CheXpert2019}.

An inherent characteristic of medical datasets is class imbalance, that is to say, the number of examples associated with one pathology is significantly larger than the number of examples associated with other pathologies.
This is due to multiple factors, including the prevalence of each pathology in the population of study or even the severity of the pathology (which could lead to multiple subsequent chest X-rays).

\begin{table}[tb]
\centering\footnotesize
\caption{Comparison of publicly available chest X-ray datasets. Here, MIMIC~\citep{MIMIC-CXR2019} refers to MIMIC-CXR-JPG~\citep{mimicjpg}.}
\renewcommand{\arraystretch}{2.5}
\begin{adjustbox}{center}
\centering\footnotesize
\begin{tabular}{@{} l r r l l l l l @{}}
\toprule
Dataset &  \#Pathol & \#Images & Period & Source  & Labelig Pipeline\\
\midrule
OpenI & 18 & 7,470 & NA & \makecell[l]{Indiana Network for Patient Care,\\Indiana, USA} & \makecell[l]{MeSH}\\
chestX-ray8 & 8 & 108,948  & \makecell[l]{1992-\\2015} & \makecell[l]{National Institutes of Health,\\USA}  & \makecell[l]{MetaMap,  DNorm,\\custom negation rules}\\
ChestX-ray14 & 14& 112,120 & \makecell[l]{1992-\\2015}& \makecell[l]{National Institutes of Health,\\USA}  & \makecell[l]{MetaMap, DNorm,\\custom negation rules}\\
CheXpert & 14 & 224,316  & \makecell[l]{2002-\\2017} & \makecell[l]{Stanford Hospital,\\California, USA} & \makecell[2l]{CheXpert}\\
MIMIC & 14 &  377,110 & \makecell[l]{2011-\\2016} & \makecell[l]{Beth Israel Deaconess Medical Center,\\Massachusetts, USA}  & \makecell[l]{CheXpert/NegBio}\\
PadChest & 19 & 168,861  & \makecell[l]{2009-\\2017}  & \makecell[l]{Hospital San Juan\\Alicante, Spain} & \makecell[l]{Physicians}\\
VinDr-CXR & 14 & 18,000   & \makecell[l]{2018-\\2020}  & \makecell[l]{Hanoi Medical University Hospital\\and Hospital 108,\\Vietnam} & \makecell[l]{VinDr Lab}\\
\bottomrule
\end{tabular}
\end{adjustbox}
\label{tbl:chest_datasets}
\end{table}

\subsection{Data}
In order to have a dataset with a more general epidemiological distribution for evaluating pathology classification models trained on a few examples, we propose MetaChest, a combination of CheXpert, MIMIC, ChestX-ray14, and PadChest, which provides a meta-learning oriented partitioning suitable for few-shot learning scenarios.
Only patients between 10 and 80 years were considered, and incomplete records as well as corrupted images were discarded.
Overall, MetaChest comprises 479,215 chest X-ray images, of which 322,475 are multi-labeled. 
Each of these images is associated with one or more of the 15 most common pathologies across the four original datasets, resulting in a total of 596,494 different pathology instances.
On the other hand, 156,740 images are normal or labeled as \textit{not finding}, indicating that no specific abnormalities were observed in the original datasets.

\begin{figure}[tb]
    \centering
    \includegraphics[width=1\textwidth, scale=1.2]{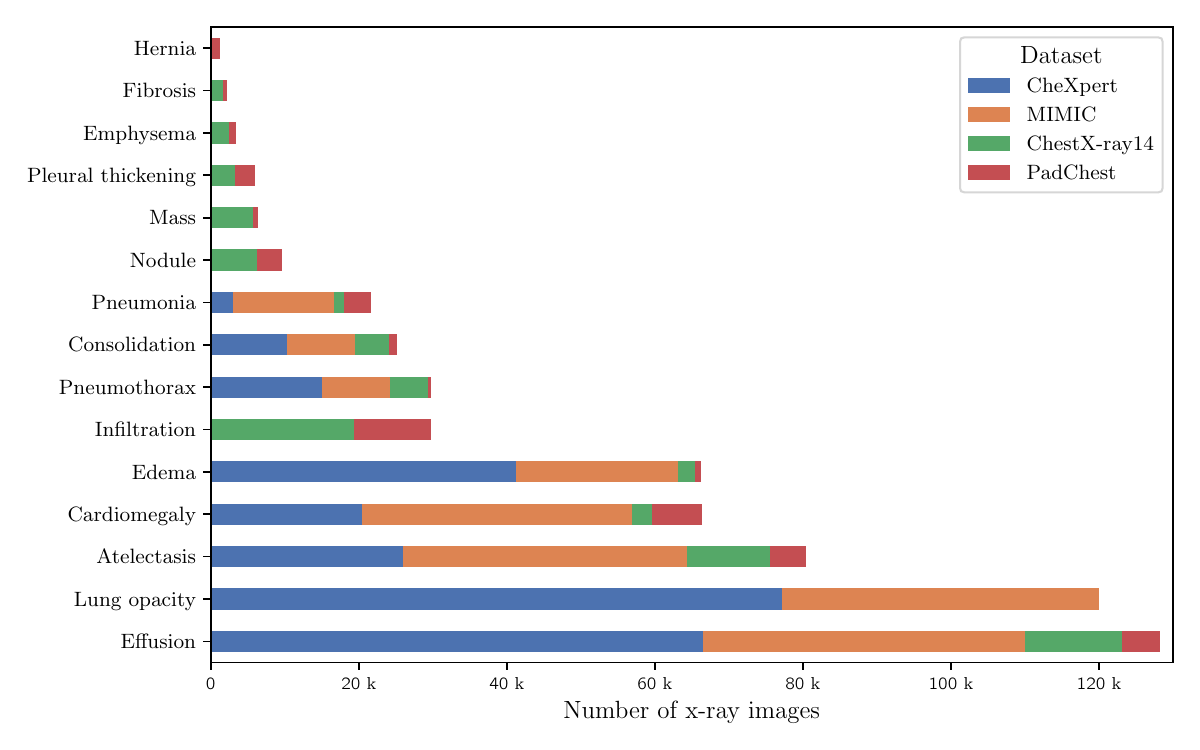}
    \caption{Distribution of labels per pathology and dataset in MetaChest.}
    \label{fig:MetaChest_distribution}
\end{figure}

The frequency of each pathology in MetaChest is shown in \autoref{fig:MetaChest_distribution}.
As can be observed, there is a pronounced class imbalance, with the most frequent pathology (Effusion) occurring nearly two orders of magnitude more often than the least frequent one (Hernia).
With respect to labeling, MetaChest has a label cardinality (average number of labels per image) of 1.84 and a label density (label over the total number of labels;  see \citet{Tsoumakas2010}) of 0.12.

\begin{figure}[tb]
    \centering
    \includegraphics[width=1\textwidth, scale=1.2,]{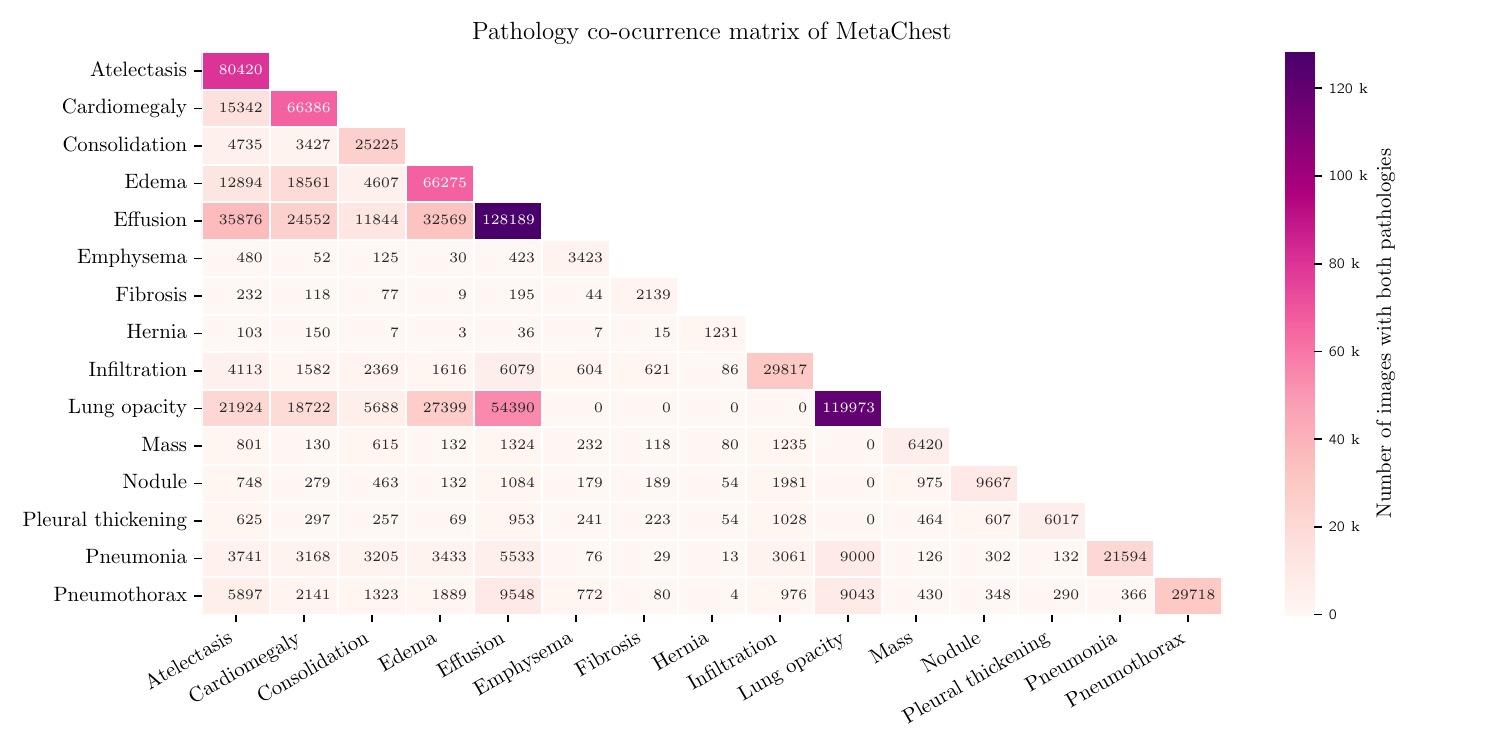}
    \caption{Co-occurrence matrix of MetaChest pathologies.}
    \label{fig:MetaChest_coocu}
\end{figure}

Label co-occurrence in MetaChest is illustrated in \cref{fig:MetaChest_coocu}. 
The most frequently co-occurring pathology pairs are Lung Opacity-Effusion, Effusion-Atelectasis and Effusion-Edema. 
Note that although Lung Opacity is the second most frequent pathology in MetaChest and frequently occurs together with five pathologies, there are seven pathologies with which it never presents together.
Moreover, Hernia is the pathology that less commonly occurs together with other pathologies, which is expected since it is also the less frequent pathology in MetaChest.

The code used to generate MetaChest are publicly available at {\url{https://github.com/bereml/metachest} and on the dataset's website at \url{https://bereml.github.io/metachest/}.

\subsection{Meta-learning partition}

In this work, we focus on pathology classification using small datasets with few classes and few examples per class.  
In particular, we consider an episode-based setting similar to SFSL, where the classifier is trained and evaluated across multiple episodes to study the model's behavior in scenarios with a small number of classes and few examples.

For this reason, we partition MetaChest classes into meta-training $\mathcal{C}_{meta-trn}$, meta-validation $\mathcal{C}_{meta-val}$, and meta-test $\mathcal{C}_{meta-tst}$ sets using the following procedure.
First, for $\mathcal{C}_{meta-tst}$ we select the five pathologies with the fewest images that are present in all four original datasets.
This allows the study of dataset shift and its impact on classification performance.
Then, from the remaining nine pathologies, we select the five with the largest number of images for $\mathcal{C}_{meta-trn}$ and the other four for $\mathcal{C}_{meta-val}$. 
Unlike the meta-test set, the meta-training and meta-validation sets contain pathologies that are not available in all four original datasets. 

\autoref{tbl:patolog} shows the classes associated with each meta-set, along with the number of examples contributed by each original dataset.
In general, CheXpert and MIMIC provide the largest number of labels for the meta-training and meta-test sets; together, these two datasets account for 77.33\% and 86.24\% of the total labels in the meta-training and meta-test sets, respectively. 
On the other hand, ChestX-ray14 and PadChest supply all the examples in the meta-validation set, as well as all the examples of several pathologies in the meta-training set.
This is due to the absence of these pathologies in CheXpert and MIMIC.
Note that although CheXpert contributes only three different pathologies (Effusion, Lung Opacity, and Atelectasis) to the meta-training set, it accounts for 44\% of the total labels in this meta-set. 
In contrast, both PadChest and ChestX-ray14 contribute six out of seven pathologies, yet cover only the 7.14\% and 15.51\%, respectively, of the total labels in the meta-training set.
Regarding normal images, there are 99,983 in the meta-training set, 1,788 in the meta-validation set, and 54,969 in the meta-test set.

\begin{table}[ht]
\caption{Meta-training, meta-validation, and meta-test class sets with the corresponding number of label instances per pathology.}
\begin{adjustbox}{center}
\centering\footnotesize
\begin{tabular}{@{}l@{\hspace{10pt}}r@{\hspace{30pt}}r@{\hspace{15pt}}r@{\hspace{15pt}}r@{\hspace{15pt}}r@{\hspace{15pt}}@{}}
\toprule
\multirow{2}{*}[-2pt]{Classes} & \multirow{2}{*}[-2pt]{\textbf{MetaChest}} & \multicolumn{4}{c}{Datasets} \\
\cmidrule(lr){3-6}
& & CheXpert & MIMIC & ChestX-ray14 & PadChest \\ 
\midrule[.2pt]
		\textit{$\mathcal{C}_{meta-trn}$} & & & & & \\
		\qquad Effusion & 128,189 & 66,484 & 43,544 & 13,086 & 5,075 \\
        \qquad Lung opacity & 119,973 & 77,194 & 42,779 &   &   \\
        \qquad Atelectasis & 80,420 & 25,980 & 38,297 & 11,335 & 4,808 \\
        \qquad Infiltration & 29,817 & & & 19,362 & 10,455 \\
        \qquad Nodule & 9,667 & & & 6,238 & 3,429 \\
        \qquad Mass & 6,420 & & & 5,682 & 738 \\
        \qquad Pleural thickening & 6,017 & & & 3,326 & 2,691 \\
        \qquad\qquad\textit{Total} & \textit{380,503} & \textit{169,658} & \textit{124,620} & \textit{59,029} & \textit{27,196} \\
        \vspace{-3mm}\\
        \textit{$\mathcal{C}_{meta-val}$} & & & & & \\
        \qquad Emphysema & 3,423 & & & 2,484 & 939 \\
        \qquad Fibrosis & 2,139 & & & 1,650 & 489 \\
        \qquad Hernia & 1,231 & & & 197 & 1,034 \\
        \qquad\qquad\textit{Total} & \textit{6,793} & & & \textit{4,331} & \textit{2,462} \\
        \textit{$\mathcal{C}_{meta-tst}$} & & & & & \\ 
        \qquad Cardiomegaly & 66,386 & 20,391 & 36,512 & 2,701 & 6,782 \\
        \qquad Edema & 66,275 & 41,247 & 21,894 & 2,269 & 865 \\
        \qquad Pneumothorax & 29,718 & 14,977 & 9,215 & 5,220 & 306 \\
        \qquad Consolidation & 25,225 & 10,340 & 9,183 & 4,505 & 1,197 \\
        \qquad Pneumonia & 21,594 & 2,986 & 13,679 & 1,381 & 3,548 \\
        \qquad\qquad\textit{Total} & \textit{209,198} & \textit{89,941} & \textit{90,483} & \textit{16,076} & \textit{12,698} \\
		\bottomrule
	\end{tabular}
\end{adjustbox}
\label{tbl:patolog}
\end{table}	

Although~\citet{cherti2022} used a chest X-ray dataset that combined multiple datasets, it is not publicly available, its generation procedure is not described, and it does not provide appropriate partitions for meta-learning evaluation.
Similarly, TorchXRayVision~\citep{Cohen2022xrv} is a library that allows the combination of different chest X-ray datasets, but it does not consider SFSL scenarios either.
Conversely, MetaChest employs a disjoint class partition, enabling experimentation in SFSL settings.
In addition, $\mathcal{C}_{meta-tst}$ is composed of the classes with the fewest examples available across the four original datasets, which is convenient for evaluating classification methods on images collected from multiple hospitals.

\section{Transfer learning strategies}
\label{sec:transfer} 

In this section, we describe the two transfer learning strategies used in this work and highlight their differences. 

\subsection{Standard transfer learning}

Standard transfer learning (STL) is the most widely studied and spread strategy for computer vision tasks. 
When performing STL, we can identify two main stages~\citep{tfsurvey}: 

\begin{itemize}
    \item Pre-training, which aims to acquire transferable knowledge from a source dataset $\mathcal{S}$.
    \item Adaptation, which leverages the knowledge acquired during pre-training to solve a task on a target dataset $\mathcal{T}$.
\end{itemize}

In the pre-training stage, $\mathcal{S}$ is divided into training $\mathcal{S}_{trn}$, validation $\mathcal{S}_{val}$ and test $\mathcal{S}_{tst}$ subsets. 
A randomly initialized neural network is then trained using batches $\mathcal{B}_{trn}$ sampled from $\mathcal{S}_{trn}$ and validated with batches $\mathcal{B}_{val}$ sampled from $\mathcal{S}_{val}$ to produce a pre-trained model.
This process is commonly repeated with different hyperparameter configurations, yielding multiple pre-trained models.
A single pre-trained model is subsequently selected based on its performance on the validation subset $\mathcal{S}_{val}$.
In some cases, the selected pre-trained model is also evaluated on the test subset $\mathcal{S}_{tst}$. 

In the adaptation stage, the target dataset $\mathcal{T}$ is typically divided into train $\mathcal{T}_{trn}$,  validation $\mathcal{T}_{val}$, and test $\mathcal{T}_{tst}$ subsets.
In order to transfer the knowledge acquired from the source dataset, a pre-trained neural network is first assembled: the feature extraction layers (also known as the backbone) are preserved with their original weights and biases, while the layers specific to the pre-training task are replaced with randomly initialized layers tailored to the target task.
Then, the assembled neural network is trained using batches $\mathcal{B}_{trn}$ sampled from $\mathcal{T}_{trn}$ and validated with batches sampled from $\mathcal{T}_{val}$ to produce the model for the target task. 
As in the pre-training stage, multiple models can be produced with different hyperparameter configurations, from which a single model is selected based on its performance on the validation subset $\mathcal{T}_{val}$. 
Finally, the performance of the selected model is estimated using the test subset $\mathcal{T}_{tst}$.

Note that the pre-training and adaptation stages in STL have some distinctive characteristics that are worth mentioning:

\begin{itemize}
    \item The classes in the source dataset $\mathcal{S}$ and the target dataset $\mathcal{T}$ are typically different; that is to say, the classes encountered during the adaptation stage were not seen during pre-training.
    \item Training is a batch-based iterative process, in which all classes within $\mathcal{S}$ are considered. 
    \item Although the target dataset is smaller than the source dataset, it typically contains examples on the order of hundreds or even thousands per category.
    \item The STL performance is evaluated on a single task $T$ that considers all classes and examples in $\mathcal{T}_{tst}$.
\end{itemize}

\subsection{Meta-learning}
\label{sec:metalearning}

Meta-learning (MTL) is an alternative paradigm to STL, which aims to obtain models that can adapt to novel tasks with unseen classes and very few examples per class~\citep{tfsurvey}.
In other words, the objective of MTL is to achieve a more efficient transfer in terms of data.  
Similar to STL, the transfer process of MTL consists of a pre-training stage followed by an adaptation stage. 
Although in MTL these stages are commonly referred to as meta-training and meta-test~\citep{optLSTM}, for the sake of consistency and clarity, we use the terms pre-training and adaptation for both STL and MTL.
In this work, we are interested in two MTL formulations for classification: standard few-shot classification (SFSC) and generalized few-shot classification.

In SFSC, the pre-training stage is equipped with the meta-training $\mathcal{D}_{meta-trn}$ and meta-validation $\mathcal{D}_{meta-val}$ datasets, while the adaptation stage uses the meta-test dataset $\mathcal{D}_{meta-tst}$.
During the pre-training stage an iterative training process is carried out. 
In each iteration, a classification task $E_{meta-trn}$ is randomly generated. This task is known as episode and is used to train the neural network.
Each episode $E_{meta-trn}$ is composed of a training $D_{trn}$ subset and a test $D_{tst}$ subset, which share the same classes. 
To generate a meta-training episode $E_{meta-trn}$, $n$ classes (known as $n$-way) are randomly selected from the set of meta-training classes $\mathcal{C}_{meta-trn}$. 
For each selected class, $k_{trn}$ and $k_{tst}$ examples are randomly sampled from $\mathcal{D}_{meta-trn}$ to form the $D_{trn}$ and $D_{tst}$ subsets.
Typically, an episode is $5$-way, and the number of samples per class are $k_{trn} = 5$ and $k_{tst} = 15$.
Once the model is trained with a meta-training episode $E_{meta-trn}$, its performance is evaluated with an episode $E_{meta-val}$ sampled from the meta-validation set $\mathcal{D}_{meta-val}$.
This pre-training process is known as episodic training. 

As opposed to STL, the adaptation stage in SFSC follows a similar iterative process as pre-training, except that the meta-test episodes $E_{meta-tst}$ are sampled from $\mathcal{D}_{meta-tst}$.
The model's performance in the adaptation stage is estimated by averaging performance scores over hundreds or thousands of episodes. 
Note that while STL focuses on assessing the capacity of the model to adapt to a single task $T$ that comprises all the examples and classes in the test subset of the target dataset, SFSC assesses the model's capacity to adapt to a large number of small episodes $E_{meta-tst}$ sampled from $\mathcal{D}_{meta-tst}$.
In other words, SFSC aims to estimate the adaptability of the model to tasks with novel classes and a few examples per class.

On the other hand, the difference between SFSC and GFSC lies in the classes and examples that constitute the $\mathcal{D}_{meta-val}$ and $\mathcal{D}_{meta-tst}$ sets.
In SFSC, the set of classes for $\mathcal{D}_{meta-val}$ ($\mathcal{D}_{meta-tst}$) is equal to $\mathcal{C}_{meta-val}$ ( $\mathcal{C}_{meta-tst}$), which is disjoint from the set of classes for $\mathcal{D}_{meta-trn}$. 
In contrast, in GFSL the set of classes for $\mathcal{D}_{meta-val}$ ($\mathcal{D}_{meta-tst}$) is equal to $\mathcal{C}_{meta-trn} \cup \mathcal{C}_{meta-val}$ ($\mathcal{C}_{meta-trn} \cup \mathcal{C}_{meta-tst}$).
Thus, GFSC can be regarded as a generalization of SFSC in which evaluation episodes are comprised not only of unseen classes sampled from $\mathcal{C}_{meta-val}$ ($\mathcal{C}_{meta-tst}$), but also seen classes from $\mathcal{C}_{meta-trn}$.

\section{Methodology}
\label{sec:episodes}

In this section, we present a formulation of few-shot multi-label classification for chest X-rays and describe a transfer learning method and a meta-learning method, which will be compared through empirical experiments.

\subsection{Few-shot multi-label classification for chest X-rays}

In this work, we focus on generalized few-shot classification, since this formulation allows modeling common medical scenarios in which one seeks to classify opacities in an X-ray image that are associated with a combination of well-known pathologies and uncommon or even novel pathologies.
Recall that in GFSC, a meta-validation or meta-test episode is composed of two types of classes.
The first type are the seen classes, which are used in the meta-training episodes during the pre-training phase.
In this sense, the seen classes are regarded as known information, even if the examples have not been previously seen.
The second type is the unseen classes, which are completely new and appear only in meta-validation episodes during pre-training or meta-testing in the adaptation stage.
These classes and examples are considered completely novel information.
The greater the number of unseen classes, the more difficult the episode, due to the higher amount of novel information, reaching a limit at the SFSC formulation (i.e., when all classes in the episode are unseen).
However, in medical scenarios, an X-ray image presents opacities that are mostly expected to be associated with known pathologies, which contrasts with SFSC, where all pathologies are unknown.  

To study this, we propose \cref{alg:episodegenerat}, which generates multi-labeled episodes and allows control over the number of seen and unseen classes, as well as the minimum number of examples per class.
Due to the multi-label nature of MetaChest, for the generation of episodes, the data is divided into $\mathcal{D}_{meta-trn}$, $\mathcal{D}_{meta-val}$, and $\mathcal{D}_{meta-tst}$ sets, as shown in \cref{fig:examplepartition}.
This division ensures that no examples are shared between meta-training, meta-validation, and meta-test episodes, making the classification task more challenging and contributing to a more robust evaluation.

\begin{figure}[tb]
  \begin{adjustbox}{center}
  \includegraphics[width=1\textwidth]{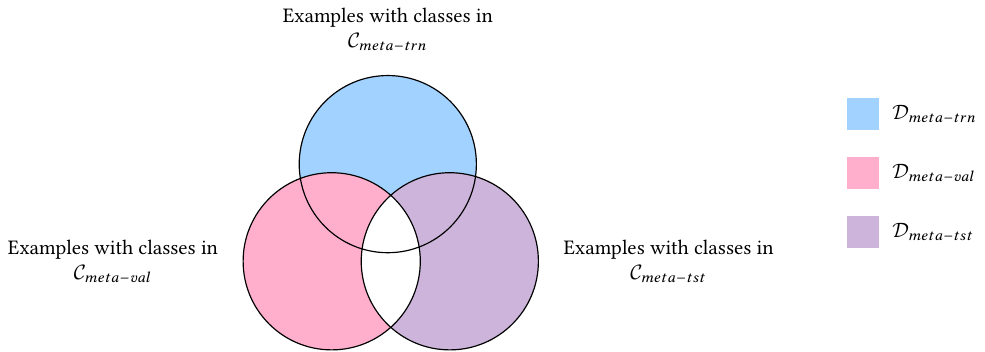}
  \end{adjustbox}
  \caption{Venn diagram illustrating the relationship between meta-training $\mathcal{C}_{meta-trn}$, meta-validation $\mathcal{C}_{meta-val}$, and meta-test $\mathcal{C}_{meta-tst}$ class sets. Blue indicates examples in $\mathcal{D}_{meta-trn}$, pink in $\mathcal{D}_{meta-val}$, and purple in $\mathcal{D}_{meta-tst}$. Since these sets are disjoint and episodes are generated from only one set at a time, the meta-validation and the meta-test episodes contain examples that are not used during meta-training.}
  \label{fig:examplepartition}
\end{figure}

The pseudocode in \cref{alg:episodegenerat} outlines the meta-validation episode generation process, which can be applied similarly to meta-test episodes.
First, we sample a set $C_{seen}$ of $n_{seen}$ classes from the meta-training classes $\mathcal{C}_{meta-trn}$, and sort them in ascending order according to their frequency in MetaChest (lines 1 and 2).
Analogously, we sample the set $C_{unseen}$ for the unseen classes (lines 3 and 4).
Given the multi-label nature of the data, we identify the set of excluded classes $C_{excluded}$ (line 5), which keeps track of the classes that do not belong to $C_{seen}$ or $C_{unseen}$. 
Then, we generate the sample set $\mathcal{D}'$ consisting of examples $x$ in MetaChest that are not labeled with any of the excluded classes $C_{excluded}$ (line 6), thereby avoiding the introduction of additional classes into the episodes.
Next, we generate the training subset $D_{trn}$ (lines 8 to 15).
For each class $c$ in ${(C_{seen} \cup C_{unseen})}$, we determine the number of missing examples $k_{missing}$ in $D_{trn}$ needed to reach $k_{trn}$ (lines 9 and 11).
After that, we sample a set $D_c$ with $k_{missing}$ examples from $\mathcal{D}'$ with class $c$ and add it to $D_{trn}$ (lines 12 and 13).
Finally, we add a \textit{not finding} X-ray example (line 15) to ensure that for every class there will be a negative example in $D_{trn}$, which allows calculation the ROC-based metrics used in this work.
The test subset $D_{tst}$ is generated in an analogous manner.

\begin{algorithm}[tb]
    \BlankLine
    \KwData{MetaChest dataset $\mathcal{D}$}
    \KwData{Sets of classes $\mathcal{C}_{meta-trn}$ and $\mathcal{C}_{meta-val}$}
    \KwData{Number of seen $n_{seen}$ and unseen $n_{unseen}$ classes}
    \KwData{Number of traning $k_{trn}$ and test $k_{tst}$ examples per class}
    \KwResult{Meta-validation episode $E_{meta-val}$}
    \BlankLine
    $C_{seen} \gets$ sample $n_{seen}$ classes from $\mathcal{C}_{meta-trn}$\;
    $C_{seen} \gets$ sort $C_{seen}$ by increasing frequency in $\mathcal{D}$\;
    $C_{unseen} \gets$ sample $n_{unseen}$ classes from $\mathcal{C}_{meta-val}$\;
    $C_{unseen} \gets$ sort $C_{unseen}$ by increasing frequency in $\mathcal{D}$\;
    $C_{excluded} \gets$ $\overline{(C_{seen} \cup C_{unseen})}$\;
    $\mathcal{D}' \gets \{x\ |\ x \in \mathcal{D} \ \text{and}\ x\ \text{is not labeled with a class}\ c \in  C_{excluded}\}$\;
    $D_{trn} \gets \emptyset$, $D_{tst} \gets \emptyset$\;
    \BlankLine
    \ForEach{$(D, k) \in \{(D_{trn}, k_{trn}), (D_{tst}, k_{tst})\}$}{
        \ForEach{$c \in C_{seen} \cup C_{unseen}$}   { 
            $k_{present} \gets |\{x\ |\ x \in D \ \text{and}\ x\ \text{is labeled with}\ c\}|$\;
            $k_{missing} \gets k - k_{present}$\; 
            $D_{c} \gets$ sample $k_{missing}$ examples labeled with $c$ from $\mathcal{D}'$\;
            $D \gets$ $D \cup D_{c}$\;
            $\mathcal{D}' \gets$ $\mathcal{D}' - D_{c}$\;
        }
    $D \gets$ $D \cup \{\text{1 \textit{not finding} image sampled from } \mathcal{D}\}$\;
    }
    $E_{meta-val} \gets (D_{trn}, D_{tst})$\;
    \BlankLine
    \KwRet{$E_{meta-val}$}
    \BlankLine
\caption{Meta-validation episode generator.}
\label{alg:episodegenerat}
\end{algorithm}

\subsection{Classification methods}
\label{sec:methods}
Let us denote the set of examples that are labeled with a class $c$ in the training episode as $D_{c} = \{(\mathbf{x}, \mathbf{y}) \ | \  (\mathbf{x}, \mathbf{y}) \in D_{trn} \ \text{and} \ \mathbf{y}[c] = 1 \}$, where $\mathbf{x} \in \mathbb{R}^{h \times w \times 3}$ is a $h \times w$ image, and $\mathbf{y} \in \{0, 1\}^{n-\text{way}}$ is the associated multi-label vector.
Furthermore, we denote $f_{\phi}(\mathbf{x}) \in \mathbb{R}^D$ as the $D$-dimensional vector representation of $\mathbf{x}$, computed by a backbone $f_{\phi}$ with trainable parameters $\phi$.

\subsubsection{ProtoNet-ML}
ProtoNet \citep{protonet} is a multi-class classification method that has been widely studied in the SFSC literature. 
In this work, we propose an extension to handle multi-label classification, which we call ProtoNet-ML. 
Following the original method, ProtoNet-ML computes a $D$-dimensional prototype $\mathbf{z}_c \in \mathbb{R}^D$ for each class $c$ as follows:

$$\mathbf{z}_c = \frac{1}{|D_{c}|} \sum_{(\mathbf{x}_i, \mathbf{y}_i) \in D_{c}}^{} f_{\phi}(\mathbf{x}_i)$$

The original multi-class ProtoNet estimates class probabilities by applying a softmax over the negative distances between a test example and the class prototypes, implicitly associating the test example to the closest prototype. 
To enable associations with multiple prototypes, ProtoNet-ML introduces a transformation function over the distances. Specifically, the transformation function $t: \mathbb{R}^D \times \mathbb{R}^D \rightarrow  \mathbb{R}$ between a test example $(\mathbf{x}, \mathbf{y}) \in D_{tst}$ and the prototype $\mathbf{z}_c$ for the class $c$ is defined as:

$$t(f_{\phi}(\mathbf{x}), \mathbf{z}_c) = \mu_c - d(f_{\phi}(\mathbf{x}), \mathbf{z}_c) $$

\noindent where $d(f_{\phi}(\mathbf{x}), \mathbf{z}_c)  = || f_{\phi}(\mathbf{x}) - \mathbf{z}_c ||$ is the Euclidean distance, and $\mu_c$ is the mean distance between the prototype for the class $c$ and all training examples in the episode, i.e., 

$$
\mu_c = \frac{1}{\vert D_{trn} \vert } \sum_{(\mathbf{x}_i, \mathbf{y}_i) \in D_{trn}} d(f_{\phi}(\mathbf{x}_{i}), \mathbf{z}_c)
$$

Subtracting the example-prototype distance from the mean distance maps examples closer than the mean to increasingly positive values, while those farther away are mapped to increasingly negative values.
This transformation can be used to compute a probability distribution for a test example $\mathbf{x}$ belonging to class $c$ as follows:

$$ p(\mathbf{y}[c] = 1 \mid \mathbf{x}) = \sigma(t(f_{\phi}(\mathbf{x}), \mathbf{z}_c))$$

\noindent where $\sigma$ denotes the sigmoid function.
Unlike multi-class prototypes, which partition the $D$-dimensional representation space into disjoint subspaces, multi-label prototypes correspond to subspaces that may overlap. 
This allows the representation of a single example to fall into more than one subspace at the same time, as shown in \cref{fig:repsubspace}.

Beyond the Euclidean distance, ProtoNet-ML can be instantiated with other functions, including the Minkowski distance and, with slight modifications, the cosine distance. 
Moreover, ProtoNet-ML is a flexible method that supports arbitrary activation functions and can operate directly on logits. In our experiments, however, we employ the sigmoid function, as it is the conventional and most natural choice for binary classification.

\begin{figure}[tb]
  \begin{adjustbox}{center}
  \includegraphics[width=.9\textwidth]{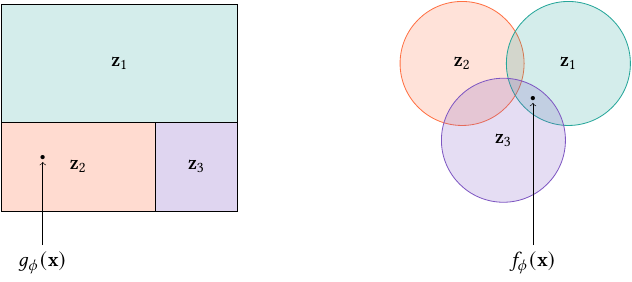}
  \end{adjustbox}
  \caption{Representation subspaces for multi-class ProtoNet (left) and ProtoNet-ML (right). In multi-class ProtoNet, a representation vector $g_{\phi}(\mathbf{x})$ is associated with only one prototype, whereas in ProtoNet-ML, a representation vector $f_{\phi}(\mathbf{x})$ may be associated with one or more prototypes. Note that $f_{\phi}(\cdot)$ is the network backbone followed by an encoding layer to reduce representation vector dimensionality.}
  \label{fig:repsubspace}
\end{figure}

\subsubsection{BatchBased}

BatchBased is a method inspired by \cite{chen2019closer} that employs STL-based training while maintaining MTL-based episode evaluation.
On top of the backbone $f_{\phi}$, BatchBased adds a head module $g_{\varphi}$ (single fully connected layer) with trainable parameters $\varphi$.
The probability distribution for a test example $\mathbf{x}$ given a class $c$ is computed as:

$$ p(\mathbf{y}[c] = 1 \mid \mathbf{x}) = \sigma(g_{\varphi}(f_{\phi}(\mathbf{x})))$$

As in STL, the learning process is carried out in epochs, during which input data is fed to the model in batches.
Note that STL batches are constructed from all classes in $\mathcal{C}_{meta-trn}$, whereas MTL episodes involve only a subset of these classes.
The model parameters $\phi$ and $\varphi$ are updated with each batch by backpropagating through the entire network. 
After the epoch is completed, an episode-based evaluation is conducted following the MTL paradigm.
Specifically, for each $E_{meta-val}$ or $E_{meta-tst}$ episode, the $f_{\phi}$ parameters are frozen, while the head module $g_{\varphi}$ is replaced and updated on $D_{trn}$.
To update the head parameters $\varphi$, an iterative process is repeated $t_{steps}$ steps.
At each step, a subset of examples $M$ is randomly sampled from $D_{trn}$, where $\vert M\vert$ is a proportion $ptc_{trn}$ of $D_{trn}$. Then, the head parameters $\varphi$ are updated with a learning rate $lr_{head}$ through backpropagation using $M$. Here, $t_{steps}$, $ptc_{trn}$, and $lr_{head}$ are considered as hyperparameters.

\section{Results and discussion}
\label{sec:experimental}

In this section, we analyze the adaptation process of different models across various formulations of the multi-label classification task using few examples per pathology.
First, we compare how two distinct learning paradigms leverage ImageNet pre-training. Next, we examine their behavior across a broad set of few-shot learning tasks designed to reflect challenges commonly encountered in medical settings. 
In addition, we investigate factors that influence the adaptation process, including image resolution, and variations in architectural connectivity patterns.
Finally, we study the influence of hyperparameters on classification performance and provide illustrative examples of the resulting model predictions.

\subsection{Experimental setup}

For empirical evaluation, certain training and method hyperparameters are fixed, while others are varied to assess their impact on classification performance.
The experimental setup is described in detail below.
The code to reproduce our main findings is publicly available at \url{https://github.com/bereml/meta-cxr}.

\paragraph{Training}

The default hyperparameter configurations are listed in \cref{tbl:hp}.
Unless otherwise specified, the reported results correspond to the BatchBased configuration.

\begin{table}[tbp]
\centering
\caption{Default hyperparameter configurations.}
\begin{adjustbox}{center}
\centering\footnotesize
\begin{tabular}{@{} l c @{}}
\toprule
Parameter & Configuration \\
\midrule
\multicolumn{2}{l}{\hspace{-5pt}\textit{Data}} \\
\hspace{10pt}Distribution & Complete \\
\hspace{10pt}Image size & 384 \\
\multicolumn{2}{l}{\hspace{-5pt}\textit{Task}} \\
\hspace{10pt}Training batch size & 64 \\
\hspace{10pt}Training $n$-way, $k_{trn}$ $k_{tst}$ & 3, 30, 30 \\
\hspace{10pt}Validation $n$-way, $k_{trn}$ $k_{tst}$ & 3, 30, 30 \\
\hspace{10pt}Test $n$-way, $k_{trn}$ $k_{tst}$ & 3, 30, 30 \\
\multicolumn{2}{l}{\hspace{-5pt}\textit{Backbone}} \\
\hspace{10pt}Architecture & MobileNetV3Small075\\
\hspace{10pt}Pre-training & I1K \\
\multicolumn{2}{l}{\hspace{-5pt}\textit{Training}} \\
\hspace{10pt}Meta-trn, meta-val, meta-tst episodes & 1,000, 100, 10,000 \\
\hspace{10pt}Max epochs & 150 \\
\hspace{10pt}Optimizer & AdamW \\
\hspace{10pt}Stop metric, patience & $HM$, 10 \\
\hspace{10pt}Float precision & 16bit \\
\multicolumn{2}{l}{\hspace{-5pt}\textit{BatchBased}} \\
\hspace{10pt}Meta-trn LR & 0.0001 \\
\hspace{10pt}Meta-val $t_{steps}$, $ptc_{trn}$, $lr_{head}$  & 100, 0.5, 0.05 \\
\hspace{10pt}Meta-tst $t_{steps}$, $ptc_{trn}$, $lr_{head}$  & 100, 0.5, 0.05 \\
\multicolumn{2}{l}{\hspace{-5pt}\textit{ProtoNet-ML}} \\
\hspace{10pt}Encoding layer type, size & Average pooling, 128 \\
\hspace{10pt}Meta-training LR & 0.0001 \\
\bottomrule
\end{tabular}
\end{adjustbox}
\label{tbl:hp}
\end{table}

\paragraph{Evaluation}

Model performance is measured as the average over 10,000 episodes sampled from the meta-test set.
As is common in GFSL \citep{gsfl}, we evaluate seen and unseen classes separately and report the harmonic mean of their scores. However, we adopt AUC–ROC instead of accuracy \citep{chestclassi,ClassificationCausal} to align with evaluation standards in the medical domain.

We employ three metrics commonly used in GFSL \citep{gsfl}, each reported with a 95\% confidence interval: one computed for the seen classes, another for the unseen classes, and a third one for the harmonic mean ($HM$), which are defined as follows.

\begin{itemize}
    \item $Seen$: The AUC-ROC of all labels of the seen classes in the episode as a single binary classification task.
    
    \item $Unseen$: The AUC-ROC of all labels of the unseen classes in the episode as a single binary classification task.
    
    \item $HM$: The harmonic mean of $Seen$ and $Unseen$, i.e.:
    $$ HM = \frac{2 \  \times \ Seen \ \times \ Unseen}{Seen \ + \ Unseen} $$
    Note that the harmonic mean is commonly used in GFSL because it mitigates the dominance of seen classes in the overall performance \citep{gsfl}.
\end{itemize}


\subsection{Leveraging ImageNet}
We begin the comparison between BatchBased and ProtoNet-ML using models that are randomly initialized and pre-trained on either ImageNet-K or ImageNet-21K. 
This experiment was performed using a MobileNetV3Large100 architecture, as it is the only pre-trained model available on both versions of ImageNet.

As shown in \cref{tbl:results_pretrain}, BatchBased consistently outperforms ProtoNet-ML across all models and metrics. 
For instance, on ImageNet-1K, BatchBased surpasses ProtoNet-ML by 4.31 $HM$ points.
When comparing the ImageNet-1K and ImageNet-21K models for BatchBased, the former achieves better results across all metrics.
For example, ImageNet-1K yields an improvement of 0.78 $HM$ points compared to ImageNet-21K.
Furthermore, 
BatchBased initialized with ImageNet-21K weights demonstrates a 4.47 $HM$ point gain over randomly initialized models.
The literature on few-example regimes in inter-domain scenarios reports inconclusive findings regarding the benefits of using pre-trained models on ImageNet-1K \citet{cherti2022}.
However, our results indicate that using pre-trained models consistently improves performance on chest X-ray images.

\begin{table}[tbh]
\caption{Comparison of randomly initialized and ImageNet-pretrained MobileNetV3Large100 models for BatchBased and ProtoNet-ML.}
\begin{adjustbox}{center}
\centering\footnotesize
\begin{tabular}{lcccc}
\toprule
Model  & $Seen$ $\uparrow$ & $Unseen$ $\uparrow$& $HM$ $\uparrow$\\
\midrule
\multicolumn{4}{l}{\hspace{-6pt}\textit{BatchBased}} \\		
Random        & 82.42$\pm$0.14 & 78.17$\pm$0.35 & 78.83$\pm$0.25 \\
ImageNet-1K   & \textbf{86.49$\pm$0.11} & \textbf{83.80$\pm$0.31} & \textbf{84.08$\pm$0.22} \\
ImageNet-21K  & 85.89$\pm$0.12 & 82.98$\pm$0.32 & 83.30$\pm$0.22 \\
\multicolumn{4}{l}{\hspace{-6pt}\textit{ProtoNet-ML}} \\	
Random       & 76.48$\pm$0.14 & 75.69$\pm$0.34 & 74.83$\pm$0.23 \\
ImageNet-1K  & 82.10$\pm$0.12 & 79.45$\pm$0.30 & 79.77$\pm$0.20 \\
ImageNet-21K & 81.89$\pm$0.12 & 80.18$\pm$0.31 & 80.06$\pm$0.21 \\
\bottomrule
\end{tabular}
\end{adjustbox}
\label{tbl:results_pretrain}
\end{table}

\subsection{Few-shot learning vs transfer learning}

Building on the results from the previous subsection, we now examine different aspects inherent to few-shot classification for BatchBased and ProtoNet-ML.
\cref{tbl:batchbased_vs_protonet} compares the results of both methods across different task configurations, while \cref{fig:bb_vs_pn} illustrates the behavioral trends of each method.

\begin{table}
\centering\footnotesize
\caption{Comparison of BatchBased and ProtoNet-ML on pathology classification tasks evaluated with $HM$.}
\begin{adjustbox}{center}
\centering\footnotesize
\begin{tabular}{c@{\hspace{4\tabcolsep}}c@{\hskip2pt}cc@{\hskip2pt}cc@{\hskip2pt}c}
\toprule
\multirow{2}{*}[-2pt]{$k$-shot} & \multicolumn{2}{c} {3-way} &\multicolumn{2}{c} {4-way} & \multicolumn{2}{c} {5-way}\\
\cmidrule(lr){2-3} \cmidrule(lr){4-5} \cmidrule(lr){6-7}
  &BatchBased & ProtoNet-ML & BatchBased & ProtoNet-ML & BatchBased &  ProtoNet-ML  \\
\midrule
\multicolumn{7}{l}{\hspace{-5pt} 1-unseen} \\
1 & 70.32$\pm$0.31 & \textbf{73.28$\pm$0.21} & 70.61$\pm$0.27 & \textbf{73.44$\pm$0.17} & 71.42$\pm$0.24 & \textbf{73.56$\pm$0.15} \\
5 & 75.63$\pm$0.29 & \textbf{79.13$\pm$0.18} & 79.23$\pm$0.20 & \textbf{79.38$\pm$0.13} & \textbf{81.41$\pm$0.15} & 79.56$\pm$0.11 \\
15 & 80.28$\pm$0.26 & \textbf{80.51$\pm$0.19} & \textbf{83.61$\pm$0.14} & 81.06$\pm$0.12 & \textbf{84.71$\pm$0.10} & 81.27$\pm$0.10 \\
30 & \textbf{82.57$\pm$0.23} & 80.47$\pm$0.20 & \textbf{84.66$\pm$0.12} & 81.06$\pm$0.12 & \textbf{85.34$\pm$0.08} & 81.24$\pm$0.10 \\\vspace{-3mm}\\
\multicolumn{7}{l}{\hspace{-5pt} 2-unseen} \\
 1 & \textbf{67.89$\pm$0.20} & 66.06$\pm$0.15 & \textbf{69.24$\pm$0.15} & 66.97$\pm$0.13 & \textbf{69.68$\pm$0.13} & 67.58$\pm$0.12 \\
 5 & \textbf{76.22$\pm$0.15} & 70.70$\pm$0.11 & \textbf{77.69$\pm$0.10} & 71.81$\pm$0.09 & \textbf{78.22$\pm$0.09} & 72.87$\pm$0.08 \\
15 & \textbf{80.20$\pm$0.14} & 71.22$\pm$0.11 & \textbf{81.26$\pm$0.09} & 72.53$\pm$0.08 & \textbf{81.48$\pm$0.07} & 73.91$\pm$0.07 \\
 30 & \textbf{81.75$\pm$0.13} & 71.15$\pm$0.11 & \textbf{82.86$\pm$0.08} & 72.58$\pm$0.08 & \textbf{82.95$\pm$0.07} & 74.01$\pm$0.07 \\\vspace{-3mm}\\
\multicolumn{7}{l}{\hspace{-5pt} 3-unseen} \\
 1 & \textbf{57.25$\pm$0.12} & 56.75$\pm$0.10 & \textbf{68.01$\pm$0.15} & 66.20$\pm$0.13 & \textbf{68.81$\pm$0.12} & 66.56$\pm$0.11 \\
5 & \textbf{65.08$\pm$0.11} & 59.59$\pm$0.09 & \textbf{75.31$\pm$0.11} & 70.89$\pm$0.09 & \textbf{76.51$\pm$0.07} & 71.52$\pm$0.07 \\
15 & \textbf{71.04$\pm$0.09} & 60.57$\pm$0.08 & \textbf{78.89$\pm$0.10} & 71.69$\pm$0.08 & \textbf{79.88$\pm$0.06} & 72.49$\pm$0.07 \\
30 & \textbf{74.02$\pm$0.08} & 60.96$\pm$0.08 & \textbf{80.59$\pm$0.10} & 71.69$\pm$0.08 & \textbf{81.51$\pm$0.06} & 72.57$\pm$0.07 \\
 \multicolumn{7}{l}{\hspace{-5pt} 4-unseen} \\\vspace{-3mm}\\
1 &  &  & 57.86$\pm$0.10 & \textbf{58.03$\pm$0.08} & \textbf{68.30$\pm$0.13} & 66.55$\pm$0.12 \\
5 &  &  & \textbf{65.13$\pm$0.08} & 61.06$\pm$0.07 & \textbf{75.12$\pm$0.09} & 71.57$\pm$0.07 \\
15 & &  & \textbf{70.38$\pm$0.07} & 62.01$\pm$0.06 & \textbf{78.51$\pm$0.08} & 72.53$\pm$0.07 \\
30 & & & \textbf{73.16$\pm$0.06} & 62.35$\pm$0.06 & \textbf{80.18$\pm$0.08} & 72.68$\pm$0.07 \\
\multicolumn{7}{l}{\hspace{-5pt} 5-unseen} \\
1 &  & & & & 58.79$\pm$0.09 & \textbf{59.34$\pm$0.07} \\
5 &  & & & & \textbf{65.73$\pm$0.07} & 62.59$\pm$0.05 \\
15 & & &  &  & \textbf{70.43$\pm$0.05} & 63.52$\pm$0.04 \\
30 & & &  &  & \textbf{73.07$\pm$0.05} & 63.87$\pm$0.04 \\
\bottomrule
\end{tabular}
\end{adjustbox}
\label{tbl:batchbased_vs_protonet}
\end{table}

\begin{figure}[tb]
  \begin{adjustbox}{center}
  \includegraphics[width=1.5\textwidth]{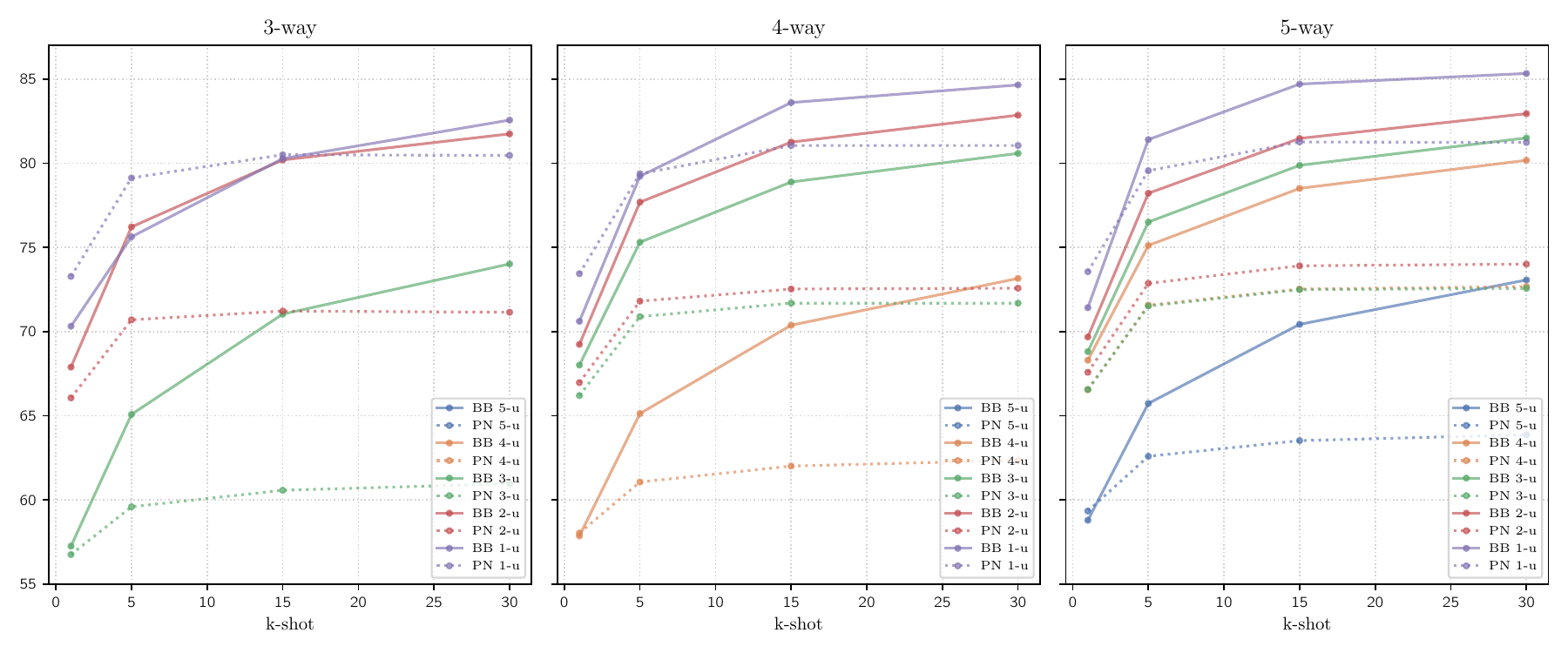}
  \end{adjustbox}
  \caption{Performance of BatchBased (BB) and ProtoNet-ML (PN) on pathology classification tasks across varying $n$-way, $n$-unseen, and $k$-shot configurations.}
  \label{fig:bb_vs_pn}
\end{figure}

We observe that ProtoNet-ML achieves improved performance in only a limited subset of task configurations. \cref{tbl:batchbased_vs_protonet} shows that these improvements occur primarily in the 1-unseen setting and tend to disappear as the number of shots ($k$-shot) or classes ($n$-way) increases.
In the remaining task configurations, BatchBased outperforms ProtoNet-ML. Moreover, as shown in \cref{fig:bb_vs_pn}, BatchBased demonstrates improved performance as the number of shots increases, whereas ProtoNet-ML’s performance remains nearly constant.
These findings are consistent with those reported by \citet{luo2023closerlookagain}, who compared the performance of conventional models with meta-learning methods such as ProtoNet on SFSC tasks across various natural image datasets. They found that conventional models tend to scale better than meta-learning approaches, particularly on fine-grained datasets.
In medical applications, where datasets often include many classes and dozens of examples per class, our results suggest that BatchBased is the more effective approach for training pathology classifiers.

\subsection{Pathology Classification Complexity}

We evaluate the effectiveness of BatchBased by varying the number of classes per episode, the number of unseen classes, and the number of examples per class. The results of these experiments are presented in \cref{tbl:k_shot}, and the corresponding performance trends are illustrated in \cref{fig:kshot_bb}.

\begin{table}[tbp]
\centering\footnotesize
\caption{$Seen$, $unseen$, and $HM$ metrics for pathology classification tasks across $n$-way, $n$-unseen, and $k_{trn}$ configurations.}
\begin{adjustbox}{center}
\addtolength{\tabcolsep}{6pt} 
\begin{tabular}{cc@{\hskip8pt}c@{\hskip8pt}cc@{\hskip8pt}c@{\hskip8pt}c}
\toprule
 \multirow{2}{*}[-2pt]{n-unseen} & \multicolumn{3}{c} {\textbf{1-shot}}  &  \multicolumn{3}{c} {\textbf{5-shot}}\\
\cmidrule(lr){2-4} \cmidrule(lr){5-7} 
& $Seen$ $\uparrow$ & $Unseen$  $\uparrow$& $HM$ $\uparrow$ &$Seen$ $\uparrow$ & $Unseen$  $\uparrow$& $HM$ $\uparrow$ \\
\midrule
\multicolumn{7}{l}{\hspace{-5pt} 3-way} \\
1 & 77.42$\pm$0.22 & 68.15$\pm$0.41 & 70.32$\pm$0.31 & 83.45$\pm$0.14 & 72.41$\pm$0.40 & 75.63$\pm$0.29 \\
2 & 80.26$\pm$0.30 & 60.70$\pm$0.20 & 67.89$\pm$0.20 & 84.18$\pm$0.21 & 70.63$\pm$0.17 & 76.22$\pm$0.15 \\
3 &              & 57.25$\pm$0.12 & 57.25$\pm$0.12 &              & 65.08$\pm$0.11 & 65.08$\pm$0.11 \\
\multicolumn{7}{l}{\hspace{-5pt} 4-way} \\
1 & 77.29$\pm$0.17 & 67.60$\pm$0.36 & 70.61$\pm$0.27 & 83.41$\pm$0.11 & 76.98$\pm$0.29 & 79.23$\pm$0.20 \\
2 & 79.13$\pm$0.17 & 62.53$\pm$0.19 & 69.24$\pm$0.15 & 84.40$\pm$0.10 & 72.47$\pm$0.15 & 77.69$\pm$0.10 \\
3 & 81.13$\pm$0.27 & 59.78$\pm$0.14 & 68.01$\pm$0.15 & 84.66$\pm$0.19 & 68.43$\pm$0.11 & 75.31$\pm$0.11 \\
4 &              & 57.86$\pm$0.10 & 57.86$\pm$0.10 &              & 65.13$\pm$0.08 & 65.13$\pm$0.08 \\
\multicolumn{7}{l}{\hspace{-5pt} 5-way} \\
1 & 77.18$\pm$0.14 & 68.65$\pm$0.34 & 71.42$\pm$0.24 & 83.26$\pm$0.10 & 80.46$\pm$0.22 & 81.41$\pm$0.15 \\
2 & 78.15$\pm$0.14 & 63.60$\pm$0.18 & 69.68$\pm$0.13 & 83.66$\pm$0.08 & 73.78$\pm$0.13 & 78.22$\pm$0.09 \\
3 & 79.84$\pm$0.16 & 61.04$\pm$0.14 & 68.81$\pm$0.12 & 84.76$\pm$0.09 & 69.99$\pm$0.10 & 76.51$\pm$0.07 \\
4 & 81.65$\pm$0.25 & 59.72$\pm$0.12 & 68.30$\pm$0.13 & 85.07$\pm$0.18 & 67.76$\pm$0.08 & 75.12$\pm$0.09 \\\vspace{2mm}
5 &              & 58.79$\pm$0.09 & 58.79$\pm$0.09 &              & 65.73$\pm$0.07 & 65.73$\pm$0.07 \\
\midrule
\multirow{2}{*}[-2pt]{n-unseen} & \multicolumn{3}{c} {\textbf{15-shot}}  &  \multicolumn{3}{c} {\textbf{30-shot}} \\
\cmidrule(lr){2-4} \cmidrule(lr){5-7} 
& $Seen$ $\uparrow$ & $Unseen$  $\uparrow$& $HM$ $\uparrow$ &$Seen$ $\uparrow$ & $Unseen$  $\uparrow$& $HM$ $\uparrow$ \\
\midrule
\multicolumn{7}{l}{\hspace{-5pt} 3-way} \\
1 & 85.08$\pm$0.12 & 78.60$\pm$0.36 & 80.28$\pm$0.26 & 85.33$\pm$0.12 & 82.13$\pm$0.32 & 82.57$\pm$0.23 \\
2 & 85.48$\pm$0.19 & 76.29$\pm$0.15 & 80.20$\pm$0.14 & 85.83$\pm$0.19 & 78.70$\pm$0.14 & 81.75$\pm$0.13 \\
3 &              & 71.04$\pm$0.09 & 71.04$\pm$0.09 &              & 74.02$\pm$0.08 & 74.02$\pm$0.08 \\
\multicolumn{7}{l}{\hspace{-5pt} 4-way} \\
1 & 85.29$\pm$0.09 & 82.71$\pm$0.21 & 83.61$\pm$0.14 & 85.60$\pm$0.08 & 84.30$\pm$0.18 & 84.66$\pm$0.12 \\
2 & 86.03$\pm$0.09 & 77.36$\pm$0.13 & 81.26$\pm$0.09 & 86.68$\pm$0.09 & 79.69$\pm$0.12 & 82.86$\pm$0.08 \\
3 & 85.90$\pm$0.18 & 73.44$\pm$0.09 & 78.89$\pm$0.10 & 86.48$\pm$0.17 & 75.90$\pm$0.08 & 80.59$\pm$0.10 \\
4 &              & 70.38$\pm$0.07 & 70.38$\pm$0.07 &              & 73.16$\pm$0.06 & 73.16$\pm$0.06 \\
\multicolumn{7}{l}{\hspace{-5pt} 5-way} \\
1 & 85.22$\pm$0.08 & 84.56$\pm$0.15 & 84.71$\pm$0.10 & 85.59$\pm$0.07 & 85.36$\pm$0.13 & 85.34$\pm$0.08 \\
2 & 85.44$\pm$0.07 & 78.13$\pm$0.11 & 81.48$\pm$0.07 & 86.21$\pm$0.06 & 80.15$\pm$0.11 & 82.95$\pm$0.07 \\
3 & 86.30$\pm$0.08 & 74.52$\pm$0.08 & 79.88$\pm$0.06 & 87.03$\pm$0.08 & 76.80$\pm$0.08 & 81.51$\pm$0.06 \\
4 & 86.34$\pm$0.16 & 72.39$\pm$0.07 & 78.51$\pm$0.08 & 86.91$\pm$0.16 & 74.79$\pm$0.06 & 80.18$\pm$0.08 \\
5 &              & 70.43$\pm$0.05 & 70.43$\pm$0.05 &              & 73.07$\pm$0.05 & 73.07$\pm$0.05 \\
\bottomrule
\end{tabular}
\end{adjustbox}
\label{tbl:k_shot}
\end{table}

\begin{figure}[tb]
  \begin{adjustbox}{center}
  \includegraphics[width=1.5\textwidth]{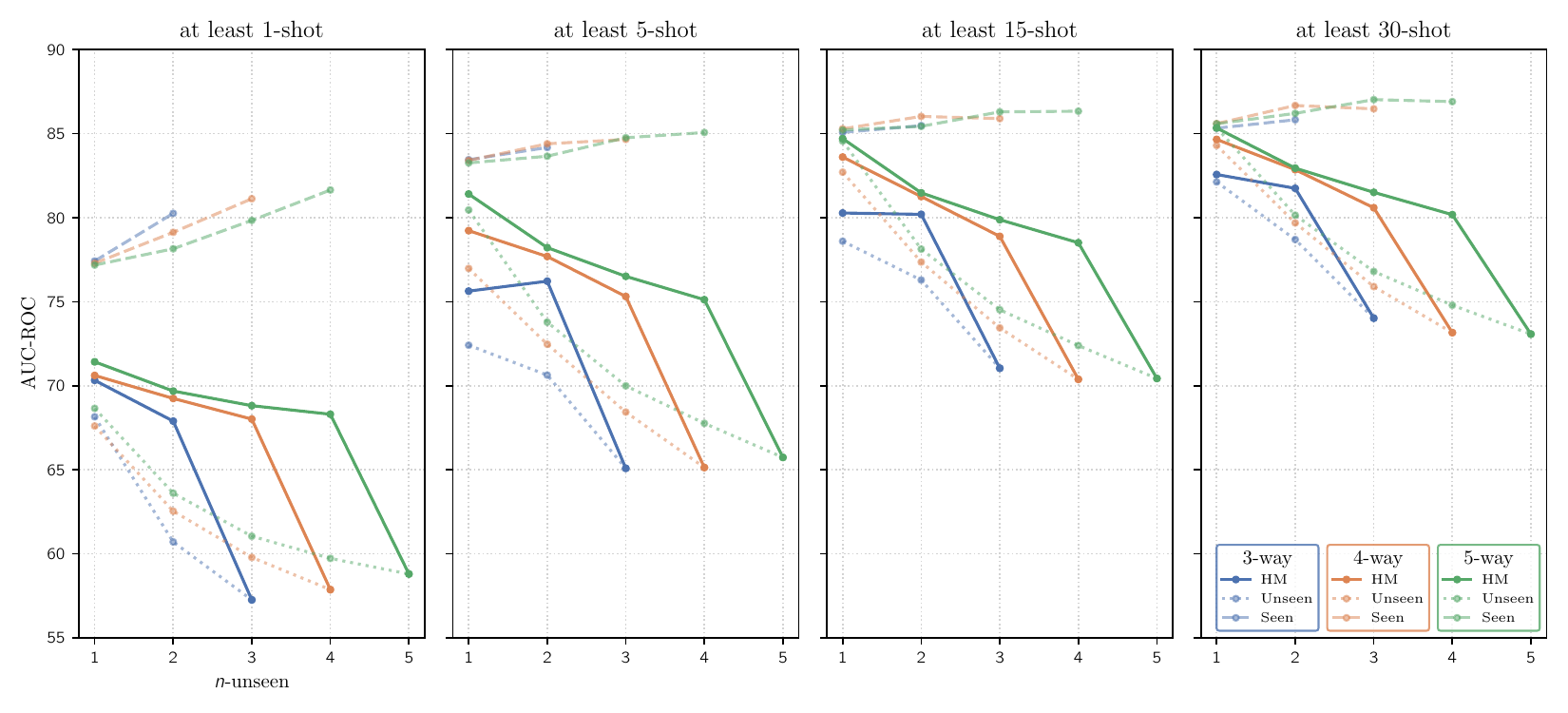}
  \end{adjustbox}
  \caption{Performance on pathology classification tasks with few examples, varying the number of classes ($n$-way) from 3 to 5, the number of unseen classes ($n$-unseen) from 1 to $n$-way, and the number of training shots per class ($k_{trn}$) to at least 5, 10, 15, and 30. Solid lines indicate the harmonic mean (HM), dashed lines indicate the AUC-ROC for seen classes, and dotted lines indicate the AUC-ROC for unseen classes.}
  \label{fig:kshot_bb}
\end{figure}

\paragraph{Classes per episode $n$-way}
We observe that, as the number of classes in the episode increases while the number of unseen classes is held constant, performance improves consistently.
For example, in \cref{tbl:k_shot}, the 5-way, 3-unseen configuration with 15-shot outperforms the 4-way, 3-unseen configuration with the same number of shots by 0.99 $HM$ points, highlighting the performance gain from adding a single class.
As shown in \cref{fig:kshot_bb}, the 4-way configurations (solid orange line) consistently achieve higher $HM$ scores than the 3-way configurations (solid blue line), while the 5-way configurations (solid green line) outperform the 4-way configurations (solid orange line).
This suggests that increasing the number of classes per episode, and consequently the number of examples, reduces task complexity and leads to improved performance.
Similar results have been reported in the SFSC literature on natural image datasets, where a higher number of classes per episode consistently improves classification performance \citep{luo2023closerlookagain}.

\paragraph{Unseen classes $n$-unseen}
We observe that performance decreases as the number of unseen classes increases, as illustrated by the downward trends in the $HM$ curves in \cref{fig:kshot_bb}.
From a learning paradigm perspective, this allows us to analyze the complexity of episodes when transitioning from a GFSL formulation (less novel information) to a standard few-shot learning SFSL formulation (entirely novel information).
Notably, the performance drop in our experiments is considerably larger when transitioning from GFSL (with at least one seen class) to SFSL (with all classes unseen).

\paragraph{Examples per class $k_{trn}$}
We observe that performance steadily improves as the number of examples per class increases.
This trend is clearly illustrated in \cref{fig:kshot_bb}, which reveals a progressive improvement in performance across the subfigures corresponding to at least 1, 5, 15, and 30 shots per class. 
For example, in~\cref{tbl:k_shot}, under the 5-way 1-unseen configuration, performance improves by 9.99, 13.29, and 13.92 $HM$ points for $k_{trn} = 5$, $k_{trn} = 15$, and $k_{trn} = 30$, respectively, compared to $k_{trn} = 1$.
This suggests that increasing the number of examples reduces task complexity, thereby enabling the model to achieve higher performance.

\paragraph{Confidence interval}
We consistently observe that increasing the number of classes results in narrower confidence intervals. Similarly, increasing the number of examples per class yields narrower intervals for both seen and unseen classes.

\subsection{X-ray resolution}

In most cases, deep neural networks used for natural image classification are trained on low-resolution images (typically $224\times224$ or $256\times256$ pixels) to reduce computational cost.
Such resolutions are adequate for datasets like ImageNet, which involve coarse-grained classification tasks characterized by visually distinct categories (e.g., cars and dogs).
Even in few-shot classification tasks on mini-ImageNet, a resolution of 64 and shallow architectures (usually 4 to 6 layers) are commonly used, helping mitigate the parameter explosion.

In contrast, classifying pathologies on chest X-rays is a fine-grained task, as the opacity patterns that distinguish different pathologies are often extremely subtle.
The literature on the effect of resolution is limited, particularly in the context of few-shot classification. Consequently, determining the most appropriate resolution for pathology classification on chest X-rays remains an important open research question.

In this experiment, we trained models using three different architectures and systematically varied the X-ray resolution to study its impact.
Images were resized using the Lanczos algorithm, a high-quality resampling method known for preserving edge sharpness and fine details~\citep{interpolationalgorithms}. The results are summarized in \cref{tbl:results_resolution}.

\begin{table}[tbh]
\caption{Comparison of MobileNetV3-Small-0.75, ConvNeXt-Tiny, and DenseNet-121 models on chest X-rays at varying input resolutions ($224\times224$, $384\times$, $512\times512$, $768\times768$, and $1024\times1024$). A fixed batch size of 32 was used to ensure a fair comparison across architectures.}
\begin{adjustbox}{center}
\centering\footnotesize
\begin{tabular}{l@{\hspace{60pt}}cccc}
\toprule
Resolution & $Seen$ $\uparrow$ & $Unseen$ $\uparrow$& $HM$ $\uparrow$\\ \midrule \multicolumn{4}{l}{\hspace{-6pt}\textit{MobileNetV3-Small-0.75}} \\			
224  & 84.29$\pm$0.13 & 81.75$\pm$0.32 & 81.87$\pm$0.23 \\
384  & 85.73$\pm$0.12 & 82.61$\pm$0.32 & 83.03$\pm$0.22 \\
512  & 85.89$\pm$0.12 & 82.53$\pm$0.32 & 83.06$\pm$0.22 \\
\textbf{768}  & \textbf{86.27$\pm$0.11} & \textbf{82.92$\pm$0.31} & \textbf{83.49$\pm$0.22} \\  
1024 & 86.39$\pm$0.11 & 82.54$\pm$0.32 & 83.23$\pm$0.23 \\ 
\multicolumn{4}{l}{\hspace{-6pt}\textit{ConvNext-Tiny}} \\	
224  & 87.22$\pm$0.11 & 84.50$\pm$0.30 & 84.88$\pm$0.21 \\
384  & 87.85$\pm$0.10 & 84.58$\pm$0.30 & 85.22$\pm$0.21 \\
512  & 88.09$\pm$0.10 & 84.44$\pm$0.30 & 85.24$\pm$0.21 \\
\textbf{768}  & \textbf{88.16$\pm$0.10} & \textbf{84.53$\pm$0.30} & \textbf{85.29$\pm$0.22} \\
\multicolumn{4}{l}{\hspace{-6pt}\textit{DenseNet-121}} \\	
224  & 84.97$\pm$0.12 & 83.27$\pm$0.29 & 83.17$\pm$0.21 \\
384  & 85.04$\pm$0.12 & 82.97$\pm$0.29 & 83.03$\pm$0.20 \\
\textbf{512}  & \textbf{85.39$\pm$0.12} & \textbf{83.37$\pm$0.29} & \textbf{83.43$\pm$0.20} \\
\bottomrule
\end{tabular}
\end{adjustbox}
\label{tbl:results_resolution}
\end{table}

All three evaluated architectures show improved performance at a resolution of $384\times384$, which is higher than the resolution commonly used in ImageNet.
For instance, MobileNetV3-Small-0.75 improves by 1.16 $HM$ points, while ConvNeXt-Tiny and DenseNet-121 achieve gains of 0.34 and 0.14 $HM$ points, respectively.

These results are consistent with previous findings in the medical imaging literature under a complete data regime.
In mammography, for example, lesions are detected more accurately in images with a resolution of $1700 \times 2100$ pixels~\citep{mammograms2018}.
Similarly, in the case of chest X-rays, \citet{chestxrayresolutions} compared the performance of two models on the ChestX-ray14 dataset using resolutions of $64\times64$ and $320\times320$ pixels, finding better performance with higher resolution.

For MobileNetV3-Small-0.75, performance progressively improves as the resolution increases up to $768\times768$, but begins to decline at higher resolutions, as shown in \cref{fig:archs}.
In contrast, both ConvNeXt-Tiny and DenseNet-121 exhibit consistent improvements with increasing resolution.
ConvNeXt-Tiny outperforms the other two architectures across all evaluated resolutions.
The highest performance is achieved with this architecture at a resolution of $768\times 768$, although it surpasses the best result of MobileNetV3-Small-0.75 by only 1.8 $HM$ points. 
ConvNeXt-Tiny improves upon MobileNetV3-Small-0.75 at the default resolution used in this work ($384\times 384$) by just 2.26 $HM$ points.

This finding is particularly relevant because increasing image resolution has an strong impact on memory requirements and computational cost for both training and inference.
In particular, this affects the memory needed for intermediate computations, gradients, and activations within the neural network, making high-resolution training substantially more demanding.
Computational cost also increases sharply, as higher resolutions require more multiply-accumulate operations (MACs) in each layer. 
In addition, the GPU memory usage grows, limiting batch sizes and potentially slowing training. 
Model complexity further interacts with image resolution: deeper or wider architectures may struggle to process very high-resolution inputs efficiently without optimization strategies such as mixed precision.
Because of these constraints, certain experiments could not be performed. 
For example, training ConvNeXt-Tiny at $1024 \times 1024$ and DenseNet-121 at $768 \times 768$ and $1024 \times 1024$ was not feasible due to GPU memory limitations and excessive computational cost, highlighting a practical limitation in scaling experiments to very high-resolution images.

\begin{figure}[tb]
    \centering
    \includegraphics[width=0.9\textwidth]{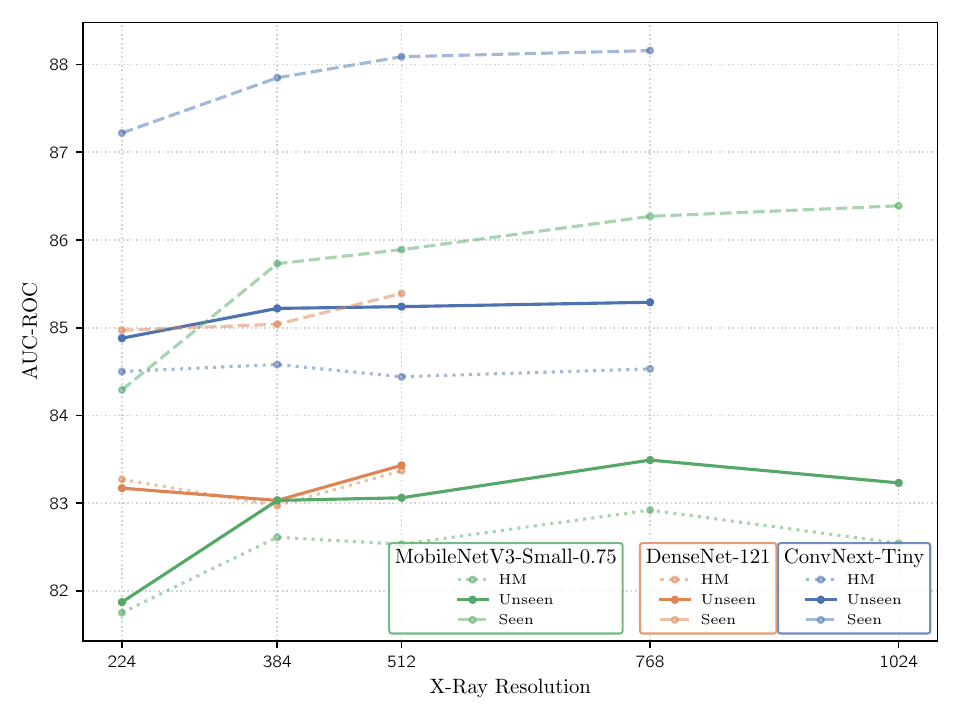}
    \caption{Comparison of convolutional architectures across progressively increasing chest X-ray resolutions.}
    \label{fig:archs}
\end{figure}

\subsection{Architectures}

We investigate how connectivity patterns and the number of parameters/operations influence pathology classification performance.
This is particularly relevant because evidence from the complete data regime in language modeling~\citep{scalinglaws, learners} and computer vision~\citep{bit2020, imageNet21Kpretraining} suggests that increasing network size and training data consistently reduces error.
For meta-learning, \citet{chen2019closer} evaluated several convolutional architectures in few-shot multi-class classification on mini-ImageNet and a reduced version of the CUB dataset~\citep{WahCUB}. 
Their findings are inconclusive: while deeper architectures improved performance on CUB, gains on mini-ImageNet were observed only in certain cases.

In this experiment, we study the effect of convolutional and attention-based connectivity patterns using popular vision architectures.
For both types, we examine efficient architectures with relatively few parameters and operations, as well as larger networks.
For efficient convolutional architectures, we focus on ConvNeXt-Atto~\citep{convnextv2} and lightweight versions of MobileNet~\citep{mobilenets}, while for Transformer-based models, we use MobileViTV2-1.0~\citep{mobilevit}.
For larger convolutional architectures, we experiment with DenseNet-121, DenseNet-161~\citep{densenet}, and ConvNeXt-Tiny, and for Transformer-based models, we consider MobileViTV2-2.0~\citep{mobilevit}. \cref{tbl:results_architectures} summarizes the results, comparing the performance of these architectures along with their number of parameters and operations.

\begin{table}[ht]
\caption{Comparison of convolutional and Transformer-based vision architectures. Models are grouped into two categories based on the number of parameters: efficient and large.}
\begin{adjustbox}{center}
\centering\footnotesize
\begin{tabular}{lcr@{\hspace{10pt}} c @{\hspace{10pt}}  c @{\hspace{10pt}}c @{\hspace{10pt}}c}
\toprule			
Architecture  & Type & Params (M)$\downarrow$ & MACs (G)$\downarrow$ & $Seen\uparrow$ & $Unseen\uparrow$ & $HM\uparrow$\\
\midrule 	
\multicolumn{4}{l}{\hspace{-6pt}\textit{Efficient}} \\
MobileNetV3-Small-075  & Conv &  1.01 &  0.11 &  85.73$\pm$0.12 & 82.61$\pm$0.32 & 83.03$\pm$0.22\\
MobileNetV3-Large-1.0  & Conv & 4.20  &  0.62 & 86.75$\pm$0.11 & 84.01$\pm$0.30 & 84.37$\pm$0.21 \\
MobileViTV2-1.0      & Tsfm & 4.38  & 4.06  & 86.13$\pm$0.11 & 82.47$\pm$0.30 & 83.21$\pm$0.21 \\
ConvNext-Atto        & Conv & 3.37 &  1.61  & 86.88$\pm$0.11 & 84.47$\pm$0.30 & 84.71$\pm$0.21 \\
\multicolumn{4}{l}{\hspace{-6pt}\textit{Large}} \\
DenseNet-121          & Conv & 6.94  &  8.09 & 85.04$\pm$0.12 & 82.97$\pm$0.29 & 83.03$\pm$0.20 \\
DenseNet-161 & Conv & 26.46 & 22.36 & 86.22$\pm$0.11 & 83.46$\pm$0.29 & 83.90$\pm$0.20 \\  
\textbf{ConvNext-Tiny} &\textbf{Conv} & \textbf{27.81} &  \textbf{18.36} &  \textbf{87.85$\pm$0.10} & \textbf{84.58$\pm$0.30} & \textbf{85.22$\pm$0.21} \\
MobileViTV2-2.0      & Tsfm & 17.42 & 16.07 &  87.15$\pm$0.11 & 84.32$\pm$0.30 & 84.75$\pm$0.21 \\
\bottomrule
\end{tabular}
\label{tbl:results_architectures}
\end{adjustbox}
\end{table}

Among the larger architectures, ConvNeXt-Tiny achieves the highest performance, reaching 85.22 $HM$ points. 
Notably, it outperforms DenseNet-161 by 1.32 $HM$ points, an architecture previously shown to be effective for medical image analysis problems~\citep{chestxrayresolutions, learningtodiagnose, pneumoniadetection, limitscohen}.

Among the efficient architectures, ConvNeXt-Atto achieves the highest performance with 84.71 $HM$ points, followed by MobileNetV3-Large-1.0. 
Compared to the default architecture in this work (MobileNetV3-Small-0.75), ConvNeXt-Atto offers an improvement of only 1.68 $HM$ points. 
However, MobileNetV3-Small-0.75 requires just 29.97\% of the parameters and 6.83\% of the computational operations used by ConvNeXt-Atto. 
This substantial reduction in resource requirements makes it particularly well-suited for deployment in resource-constrained environments, such as on-device medical image analysis systems.

Interestingly, across both efficient and larger architectures, convolutional models outperform their Transformer-based counterparts. This trend is illustrated in \cref{fig:params-flops}, which depicts the relationship between model performance and computational efficiency for the evaluated architectures.

\begin{figure}[tb]
  \begin{adjustbox}{center}
  \includegraphics[width=1.3\textwidth]{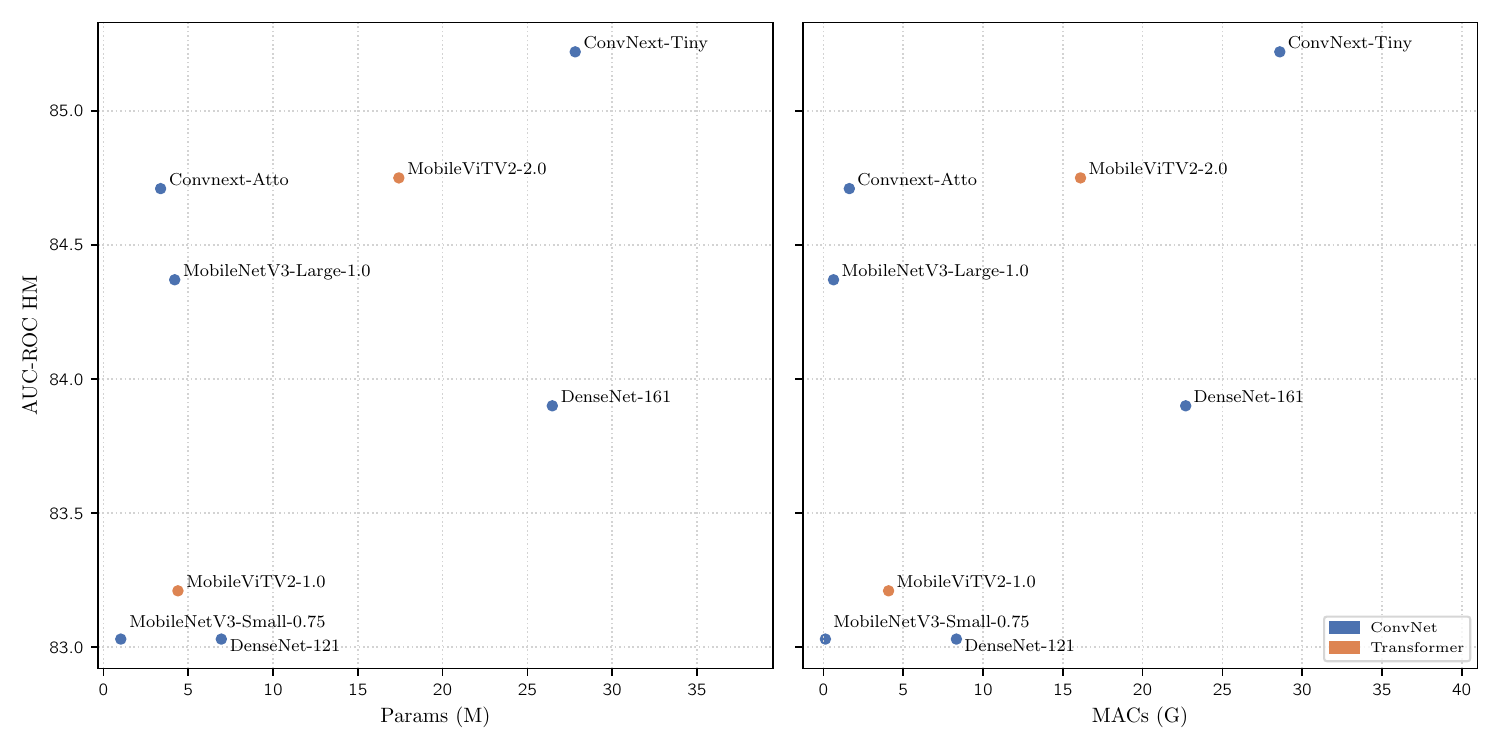}
  \end{adjustbox}
  \caption{Trade-off between classification performance and the number of parameters or multiply-accumulate operations (MACs) for different convolutional and Transformer architectures.}
  \label{fig:params-flops}
\end{figure}

Efficient neural network architectures reduce computational requirements without significantly compromising performance, offering several practical benefits. 
These architectures can run on devices with limited hardware resources, lowering costs and expanding accessibility. 
Moreover, their efficiency could enable scalable deployment and seamless integration into existing medical infrastructures, including those in remote or resource-constrained regions.


\subsection{Hyperparameter analysis}

We assess the impact of different hyperparameter configurations on method performance by varying selected hyperparameters and evaluating the resulting classification outcomes.
For the BatchBased method, two hyperparameters are tuned during the adaptation phase. The first is the Meta-tst learning rate $lr_{head}$ used to update the head parameters $\varphi$. The second is the proportion of Meta-tst $ptc_{trn}$ examples incorporated into the training steps of each episode $E_{meta-tst}$.
In the case of ProtoNet-ML, we investigate two types of encoding layer across different output sizes: a fully connected layer and an average pooling layer. The results of these experiments are shown in \cref{tbl:hyperparameter}.

For BatchBased, the lower learning rate of $0.005$ consistently yields the best results, regardless of the chosen $ptc_{trn}$ value. Among the three evaluated configurations of $ptc_{trn}$, the highest performance is achieved with a value of $0.5$. In contrast, ProtoNet-ML obtains its best performance when using an average pooling layer with an output size of $128$. Overall, BatchBased shows low sensitivity to hyperparameter variations, maintaining consistent performance across configurations and thus underscoring its robustness.

\begin{table}[ht]
\caption{Comparison of model performance across BatchBased and ProtoNet-ML hyperparameter configurations.}
\begin{adjustbox}{center}
\centering\footnotesize
\begin{tabular}{l@{\hspace{80pt}}lccc}
\toprule
Hyperparameter  & $Seen$ $\uparrow$ & $Unseen$ $\uparrow$& $HM$ $\uparrow$\\
\midrule
\multicolumn{4}{l}{\hspace{-5pt}\textit{BatchBased}} \\
\multicolumn{4}{l}{\hspace{3pt}\textit{Meta-tst  $lr_{head}=0.01$, $ptc_{trn}=$}} \\		
\hspace{11pt}0.25  & 85.07$\pm$0.12 & 81.50$\pm$0.33 & 82.01$\pm$0.24 \\
\hspace{11pt}0.5   & 85.12$\pm$0.12 & 81.55$\pm$0.33 & 82.07$\pm$0.23 \\
\hspace{11pt}0.75  & 85.11$\pm$0.12 & 81.55$\pm$0.33 & 82.07$\pm$0.23 \\
\multicolumn{4}{l}{\hspace{3pt}\textit{Meta-tst  $lr_{head}=0.005$, $ptc_{trn}=$}} \\		
\hspace{11pt}0.25  & 85.34$\pm$0.12 & 82.15$\pm$0.32 & 82.54$\pm$0.23 \\
\hspace{11pt}0.5   & \textbf{85.33$\pm$0.12} & \textbf{82.13$\pm$0.32} & \textbf{82.57$\pm$0.23} \\
\hspace{11pt}0.75  & 85.32$\pm$0.12 & 82.16$\pm$0.32 & 82.54$\pm$0.23 \\
\multicolumn{4}{l}{\hspace{-5pt}\textit{ProtoNet-ML}} \\
\multicolumn{4}{l}{\hspace{3pt}\textit{Average Pooling}} \\	
\hspace{11pt}96  & 80.16$\pm$0.14 & 79.02$\pm$0.36 & 78.17$\pm$0.25 \\
\hspace{11pt}128 & \textbf{81.88$\pm$0.12} & \textbf{80.95$\pm$0.30} & \textbf{80.47$\pm$0.20} \\
\hspace{11pt}144 & 80.61$\pm$0.15 & 77.90$\pm$0.37 & 77.70$\pm$0.27 \\ 
\multicolumn{4}{l}{\hspace{3pt}\textit{Fully Connected Layer}} \\	
\hspace{11pt}96  & 80.81$\pm$0.14 & 79.06$\pm$0.36 & 78.51$\pm$0.25 \\
\hspace{11pt}128 & 82.05$\pm$0.15 & 76.20$\pm$0.38 & 77.44$\pm$0.27 \\
\hspace{11pt}144 & 81.12$\pm$0.14 & 77.46$\pm$0.37 & 77.76$\pm$0.26 \\
\bottomrule
\end{tabular}
\end{adjustbox}
\label{tbl:hyperparameter}
\end{table}


\subsection{Visualization of model predictions}
\label{sec:visualization}

We visualize a set of model predictions to further examine its behavior qualitatively.
\cref{fig:visualization} presents selected chest X-ray examples along with their corresponding predictions across the four datasets that comprise MetaChest. 
The examples are arranged from left to right, progressing from correctly classified cases to those exhibiting substantial errors.
For example, in the last row, the image in column (\textit{a}) shows a PadChest X-ray for which the model correctly predicts all four seen classes as well as the unseen class. In contrast, the image in column (\textit{d}) of the same row illustrates a case in which the model correctly identifies three categories but misclassifies two pathologies, one seen and one unseen, both in red.

\begin{figure}[tbp]
  \begin{adjustbox}{center}
  \includegraphics[width=1.2\textwidth]{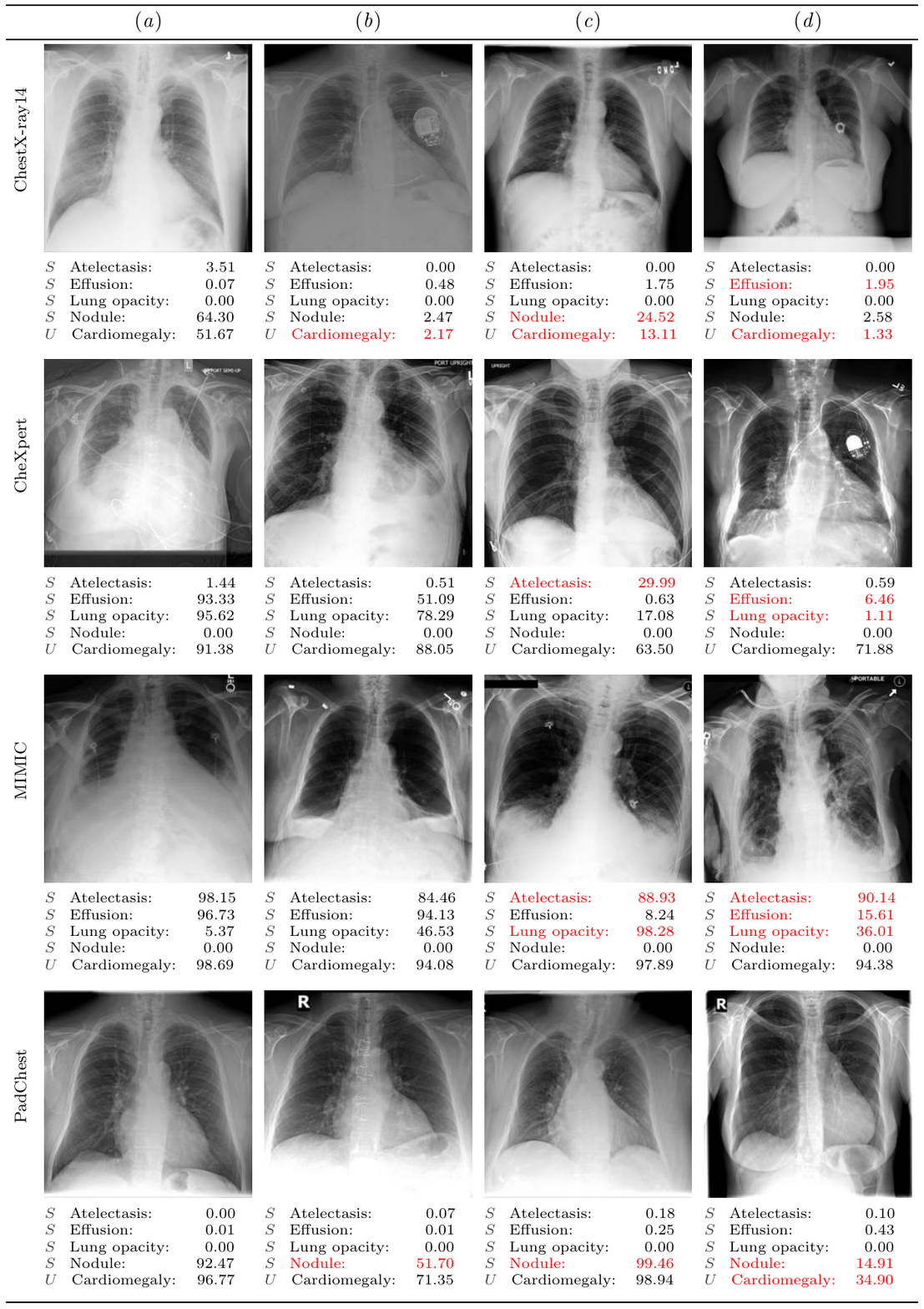}
  \end{adjustbox}
   \caption{X-ray examples with predicted labels for the 5-way, 1-unseen classification task with 30-shots per class. Each row shows four examples per dataset in MetaChest. Labels below each chest X-ray are annotated with $S$ and $U$ for seen and unseen classes, respectively. Numbers indicate predicted class probabilities; incorrect predictions are shown in red.}
  \label{fig:visualization}
\end{figure}

\section{Conclusions and future work}
\label{sec:conclu}
In this work, we investigated the key factors that influence model training for pathology classification in chest X-rays under few-shot scenarios.
To this end, we introduced MetaChest, a benchmark integrating four publicly available chest X-ray datasets. MetaChest provides a meta-set data partition specifically designed for standard few-shot classification, along with a novel multi-label episode generation algorithm.
Using MetaChest, we generated diverse classification tasks to compare two representative learning methods: one based on standard transfer learning and another widely adopted in standard few-shot classification.
We further analyzed how various factors contributing to task complexity impact model performance, including the number of examples per class ($k_{trn}$), the number of classes per episode ($n$-way), and the number of unseen classes ($n$-unseen). Additionally, we explored the effects of image resolution, as well as the connectivity patterns and computational requirements associated with each of the evaluated architectures.

The adoption of the generalized few-shot learning paradigm aligns more closely with the clinical presentation of pathologies in chest X-rays than the standard few-shot classification paradigm. 
This task formulation is particularly well-suited to specialized medical contexts and useful for a variety of scenarios within the healthcare domain.
In addition, the proposed multi-label episode generation algorithm enables the creation of complex classification tasks, further broadening its applicability to real-world medical settings.
Interestingly, our results show that BatchBased is an effective classification method in few-shot scenarios, despite being based on standard transfer learning and not specifically designed for few-shot learning. 
We also observed that increasing the number of classes per episode ($n$-way) and the number of training examples per class ($k_{trn}$) improves model performance by enhancing task robustness.
With respect to image resolution, we found that using higher resolutions than those commonly applied in natural image tasks leads to better classification performance. 
This improvement is likely due to the fine-grained nature of pathology classification, where abnormal patterns are subtle and can be overlooked at lower resolutions. 
However, this performance improvement comes at the cost of higher computational demands and longer training and inference times. 
In contrast, our results show that efficient architectures can achieve performance comparable to larger models while substantially reducing computational overhead. 
This is particularly advantageous in resource-constrained environments, such as remote areas or small hospitals, where these architectures strike a balance between performance, computational efficiency, and practical deployability. 

As future work, we envision four main research directions.
First, leveraging Vision Foundation Models as a starting point for pathology classification. 
Pre-trained on large-scale datasets, these models could provide richer and more generalizable feature representations, thereby enhancing classification performance.
Second, developing multimodal classification models that integrate complementary information from radiology reports, such as radiologist notes and clinical records. 
Incorporating this additional contextual information could enrich the diagnostic process and improve overall performance.
Third, analyzing the behavior of ProtoNet-ML under different distance and activation functions.
Finally, conducting a comparative evaluation between model predictions and expert radiologist assessments. 
Such a study would enable clinical validation of the results and provide a more accurate and reliable measure of the model’s effectiveness in clinical settings.

\subsection*{Acknowledgements}

We would like to thank Ricardo Montalvo Lezama for his valuable insights and support in the development of this work. We also extend our heartfelt thanks to Pasita and Chinito for their unwavering support throughout this process.

\bibliography{references}
\end{document}